\documentclass[10pt,twocolumn,letterpaper]{article}

\usepackage{cvpr}              % To produce the CAMERA-READY version
\newcommand{\vx}{\mathbf{x}}
\newcommand{\vz}{\mathbf{z}}
\newcommand{\vh}{\mathbf{h}}

\newcommand{\Ours}{\texttt{BayesMM}}

\newcommand{\tableCellHeight}{1}
\newcommand{\tabstyle}[1]{
  \setlength{\tabcolsep}{#1}
  \renewcommand{\arraystretch}{\tableCellHeight}
  \centering
  \small
}

\usepackage[export]{adjustbox}

\usepackage{graphicx}
\usepackage{booktabs} % for professional tables
\usepackage{multirow}  % for multirow cells in tables
\usepackage{amsmath} 
\usepackage{amssymb}
\usepackage{mathtools}
\usepackage{amsthm}
\usepackage{pifont}
\usepackage{color}
\usepackage{bbm}
\usepackage{array} 
\usepackage{makecell}
\usepackage{bm}
\usepackage{algorithm}
\usepackage{algorithmic}
\usepackage{placeins}

\definecolor{cvprblue}{rgb}{0.21,0.49,0.74}
\usepackage[pagebackref,breaklinks,colorlinks,allcolors=cvprblue]{hyperref}
\usepackage{multirow}
\def\paperID{*****} % *** Enter the Paper ID here
\def\confName{CVPR}
\def\confYear{2026}

\title{Adapting Point Cloud Analysis via Multimodal Bayesian Distribution Learning}

\author{
Xingyu Zhu\textsuperscript{1},
Liang Yi\textsuperscript{1},
Shuo Wang\textsuperscript{1},
Wenbo Zhu\textsuperscript{2},
Yonglinag Wu\textsuperscript{3},\\
Beier Zhu\textsuperscript{1*},
Hanwang Zhang\textsuperscript{4} \\[0.5em]
\textsuperscript{1}MoE Key Lab of BIPC, University of Science and Technology of China, \\
\textsuperscript{2}Opus AI Research, \
\textsuperscript{3}Southeast University,  \\
\textsuperscript{4}Nanyang Technological University, \\
{\tt\small xyzhuxyz@mail.ustc.edu.cn}
}

\begin{document}
\maketitle
\begin{abstract}
Large multimodal 3D vision--language models show strong generalization across diverse 3D tasks, but their performance still degrades notably under domain shifts. This has motivated recent studies on test-time adaptation (TTA), which enables models to adapt online using test-time data. Among existing TTA methods, cache-based mechanisms are widely adopted for leveraging previously observed samples in online prediction refinement. However, they store only limited historical information, leading to progressive information loss as the test stream evolves. In addition, their prediction logits are fused heuristically, making adaptation unstable.
To address these limitations, we propose \texttt{BayesMM}, a Multimodal Bayesian Distribution Learning framework for test-time point cloud analysis. BayesMM models textual priors and streaming visual features of each class as Gaussian distributions: textual parameters are derived from semantic prompts, while visual parameters are updated online with arriving samples. The two modalities are fused via Bayesian model averaging, which automatically adjusts their contributions based on posterior evidence, yielding a unified prediction that adapts continually to evolving test-time data without training.
Extensive experiments on multiple point cloud benchmarks demonstrate that BayesMM maintains robustness under distributional shifts, yielding over 4\% average improvement.

\end{abstract}
    
\let\thefootnote\relax\footnote{*Corresponding author.}
\section{Introduction}
\label{sec:intro}
3D sensors such as LiDAR and RGB-D cameras~\cite{FanXWWZ21, GuoWHLLB21} 
have become fundamental to robotics and autonomous driving 
for their reliable geometric perception~\cite{LuXWXTKZ23,WangCGHCW19}, 
driving advances in scene reconstruction, object recognition, and spatial understanding. 
Building on this foundation, large multimodal 3D models~\cite{xue23ulip,xue24ulip2,0006KWCY0024} 
have recently emerged by leveraging contrastive pre-training on large-scale point–image–text triplets. 
They align geometric and textual representations within a shared embedding space, 
enabling open-vocabulary point cloud recognition and strong zero-shot generalization~\cite{liu23openshape, zhou24uni3d, ZhuZ00HZ24, abs-2510-24038}, 
thus demonstrating the potential of multimodal learning for scalable and generalizable 3D perception~\cite{DeitkeLWNMKFLVG23, detector,wu2025generalization,wu2025number}.

\begin{figure}
  \centering
    \includegraphics[width=1\linewidth]{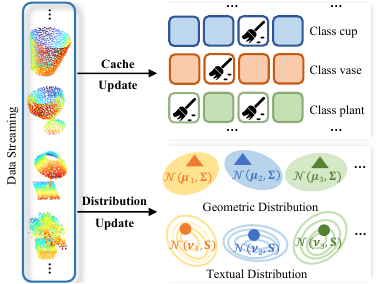}
    \caption{Comparison between cache-based adaptation and our distribution-based modeling. 
(a) Cache-based methods rely on discrete memory updates, storing a small number of recent samples in a fixed-size cache.  
(b) Our \texttt{BayesMM} models class-wise distributions across modalities rather than individual samples.
}
    \label{fig:motivation}
\end{figure}
Despite the remarkable progress of large multimodal 3D models, 
their performance often degrades when facing domain shifts between training and testing distributions. 
Recent studies have thus explored \textit{test-time adaptation (TTA)}~\cite{tent, TDA, Point-Cache, Multi-Cache, choi22improving, promttuning, zhang22memo, protomm}, enabling models to refine predictions dynamically using unlabeled test data without retraining.
Among these approaches, cache-based methods have shown particular promise by maintaining a compact memory of high-confidence test samples for online model adjustment. 
However, their limited cache capacity causes progressive information loss, 
failing to capture long-term distributional statistics, as illustrated in the top of Figure~\ref{fig:motivation}. 
As the test stream evolves, continuous sample replacement further amplifies this problem, 
leading to unstable adaptation and even catastrophic forgetting. 
In addition, the heuristic fusion between cache-based and zero-shot logits 
relies on empirically tuned hyperparameters~\cite{DMN, TDA, Point-Cache}, 
making the adaptation process unstable across domains.
These limitations hinder the practicality of cache-based methods in real-world scenarios.

To address the above limitations, we propose \texttt{BayesMM}, 
a training-free dynamic Bayesian distribution learning framework for adaptive point cloud recognition. 
As illustrated in the bottom part of Figure~\ref{fig:motivation}, 
\texttt{BayesMM} jointly models textual and geometric modalities 
under a unified probabilistic formulation. 
Specifically, it assumes that features of each class follow Gaussian distributions in both modalities. 
The textual distributions are first derived from semantic embeddings, 
providing class-wise priors that capture
semantic diversity across prompt variants.
while the geometric distributions are progressively refined from the test stream to reflect visual variations.
The Bayesian formulation allows the model to automatically adjust modality weights, yielding a unified predictive distribution that ensures stable and consistent adaptation.

\begin{figure}
  \centering
    \includegraphics[width=1\linewidth]{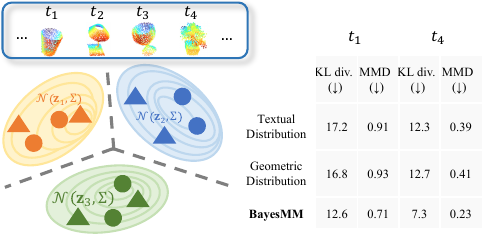}
    \caption{ Comparison of distribution consistency across adaptation steps. 
The Kullback–Leibler (KL) divergence and Maximum Mean Discrepancy (MMD) 
are measured at different time steps during test-time adaptation. 
}
    \label{fig:motivation2}
\end{figure}

To quantify the multimodal consistency achieved by \textit{BayesMM}, 
we measure the Kullback–Leibler (KL) divergence~\cite{KL} and Maximum Mean Discrepancy (MMD)~\cite{MMD}
between the learned multimodal distributions and their ground-truth references along the adaptation trajectory. 
As shown in the right part of Figure~\ref{fig:motivation2}, 
the full Bayesian fusion attains substantially lower KL and MMD values than its single-modality ablations, 
indicating more coherent alignment between textual and geometric representations. 
Both metrics steadily decrease as adaptation proceeds, 
showing that \texttt{BayesMM} continuously refines the joint feature space rather than overfitting to short-term samples. 
Specifically, the average KL divergence drops from 17.2 to 12.6, 
and the MMD decreases from 0.91 to 0.71 between the initial and later stages ($t_1$–$t_4$), 
demonstrating that our Bayesian fusion effectively stabilizes feature dynamics 
and enhances cross-modal distribution consistency over time.\footnote{$t_1$ and $t_4$ correspond to the 500\textsuperscript{th} and 2000\textsuperscript{th} test samples on ModelNet-C~\cite{ModelNet-C}, respectively.}

The main contributions are summarized as follows:
\begin{itemize}
    \item We propose \texttt{BayesMM}, a training-free dynamic Bayesian distribution learning framework for adaptive point cloud recognition at test time. 
    \item We formulate a unified probabilistic model that jointly models geometric and textual modalities through dynamic parameter updates and Bayesian fusion to achieve cross-modal alignment. 
    \item We conduct extensive experiments on multiple benchmarks, demonstrating that \texttt{BayesMM} achieves robust and consistent test-time adaptation performance across diverse 3D scenarios.
\end{itemize}

\section{Related Work}

\noindent{\textbf{Test-time adaptation with large multimodal 3D models.}} 
TTA~\cite{dota,TDA, protomm} aims to address distribution shifts by adapting model representations during inference using test data, without accessing source data.
In the 3D domain, recent works such as MATE~\cite{MATE}, BFTT3D~\cite{BFTT3D}, and CloudFixer~\cite{Cloudfixer} explore adaptive strategies for point cloud recognition through masked auto-encoding, prototype memory, and diffusion-based restoration.
However, these approaches depend on source-domain data, making them less suitable for TTA.
The emergence of large multimodal 3D models~\cite{F-VLM,YuHDSC23,PointCLIP,PointCLIPV2} has enabled generalizable and open-vocabulary 3D understanding.
Representative models such as ULIP-2~\cite{xue24ulip2}, OpenShape~\cite{liu23openshape}, and Uni3D~\cite{zhou24uni3d} jointly pre-train on large-scale point–image–text triplets via contrastive alignment~\cite{ChenK0H20}, unifying geometric and semantic representations for zero-shot generalization.
Building on these foundation models, TTA in large multimodal 3D models has recently been explored through {cache-based mechanisms}~\cite{cachemodel,Tipadpter,Point-Cache,TDA,wu2025kris,wu2025video}.
These methods maintain a cache of feature representations collected during inference and retrieve relevant entries to guide prediction updates, enabling efficient on-the-fly adaptation.
In contrast, our \texttt{BayesMM} formulates test-time adaptation as {dynamic multimodal distribution learning}.
It continuously models geometric distributions and integrates them with textual priors via Bayesian inference, where modality weights are automatically adjusted under the Bayesian principle, enabling robust cloud recognition under distribution shifts.

\noindent{\textbf{Distribution learning.}}
Distribution learning provides a principled framework for adapting recognition models by exploiting the statistical structure of feature space rather than relying on fixed representations.
Classical approaches such as Gaussian Discriminant Analysis~\cite{WangLS0WT24,hastie1996discriminant, ZhuZ00HZ24} assume that features of each class follow Gaussian distributions and construct probabilistic classifiers in closed form.
Recent advances extend this idea to test-time scenarios.
DOTA~\cite{dota} formulates test-time adaptation of vision–language models as online estimation of Gaussian parameters from test data streams to capture non-stationary shifts.
Online Gaussian Test-Time Adaptation~\cite{ZanellaFVA25,FuchsZV25} further updates class-wise means and covariances during inference for continual adaptation without external memory.
More recently, ADAPT~\cite{abs-2508-15568} aligns Gaussian-distributed test features with class prototypes by adjusting per-class statistics, while BCA~\cite{Zhou0LLZDLL25} incrementally refines Gaussian parameters using incoming samples for efficient and stable source-free adaptation.
Different from these unimodal Gaussian-based approaches, \texttt{BayesMM} performs {multimodal distribution learning} by jointly estimating geometric and textual distributions and integrating them through Bayesian inference for robust point cloud recognition.

\begin{figure*}[t]
  \centering
    \includegraphics[width=1\linewidth]{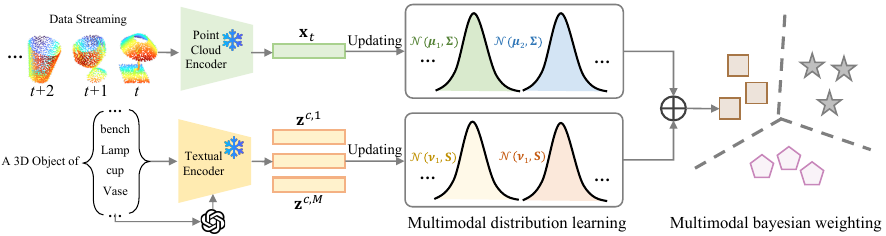}
    \caption{Overview of the proposed \texttt{BayesMM} framework. A frozen point cloud encoder extracts geometric features from streaming inputs, 
and a frozen language model provides textual embeddings as semantic priors. 
Both modalities are represented by Gaussian distributions, 
where geometric ones are updated online with incoming samples. 
Bayesian weighting fuses the two modalities into a unified posterior 
for adaptive and training-free point cloud recognition.
}
    \label{fig:framework}
\end{figure*}
\section{Methodology}
% This section elaborates on the details of Point-Cache,  
% the overall pipeline of our approach is depicted in Fig.~\ref{fig:architecture}.
In this section, we present \texttt{BayesMM}, a test-time adaptation framework that models geometric and textual modalities as evolving distributions for robust 3D recognition. 
An overview is shown in Figure~\ref{fig:framework}.
\subsection{Setup}
We consider a streaming test-time scenario for large multimodal 3D data~\cite{PointCLIP,PointCLIPV2,Point-Cache,xue23ulip,xue24ulip2}, 
where a sequence of point clouds $\{X_t\}_{t=1}^\infty$ arrives online, accompanied by a fixed set of text prototypes 
$\{T_c\}_{c=1}^C$ (\eg, ``a 3D object of $\texttt{[classname]}_c$''). 
A \emph{frozen} point encoder $\Phi$ and a \emph{frozen} text encoder $\Psi$ project the inputs into a shared feature space:
\[
\vx_t = \Phi(X_t)\in\mathbb{R}^d, \qquad \vz_c = \Psi(T_c)\in\mathbb{R}^d.
\]
On top of these fixed embeddings, a lightweight head 
$f_{\theta_t}:\mathbb{R}^d\!\to\!\mathbb{R}^C$ produces  prediction scores, 
where the parameters $\theta_t$ are updated online as new test samples arrive.

As an example, we illustrate the cache-based test-time adaptation strategy~\cite{Point-Cache}.
At the initial time ($t=0$), the classifier reduces to the zero-shot one, whose parameters are given by the text prototypes
$\theta_0=\{\vz_c\}_{c=1}^C$.
For a sample $\vx_0$, the class score is computed as:
\begin{equation}
f_{\theta_0}(\vx_0)_c = \vz_c^\top \vx_0.
\end{equation}
At time step $t$, the model maintains a class-wise cache $\vh_{t,c}$ that stores up to $K$ historical embeddings of test samples predicted with high confidence as class $c$.
% stores the historical embeddings of test samples
% predicted with high confidence as class $c$.
The parameters at time $t$ are thus
$\theta_t = \{\vz_c, \vh_{t,c}\}_{c=1}^C$.
Given a new test sample $\vx_t$, the scoring function combines the text similarity and the cache similarity:
\begin{equation}
\label{eq:cache-score}
f_{\theta_t}(\vx_t)_c
= \vz_c^\top \vx_t
+ \lambda \exp(-\gamma[1 - \mathrm{cos}(\vx_t, \vh_{t,c})]),
\end{equation}
where $\lambda>0$ balances the contributions of the zero-shot prototype and the cached features, and $\gamma$ controls the sensitivity of cosine distance.

\subsection{Multimodal distribution learning}
Cache-based adaptation~\cite{Point-Cache, TDA, DMN} suffers from two issues: limited cache capacity causes {information decay}, and heuristic logit fusion based on empirical hyperparameters (e.g., $\lambda$, $\gamma$ in Eq.~\eqref{eq:cache-score}) lacks a theoretical principle. 
In contrast, our \texttt{BayesMM} models textual and geometric modalities as {distributional representations} and fuses their classification results under a {Bayesian model averaging} formulation, effectively utilizing information from previous samples during continuous updates.

\noindent{\textbf{{Textual distribution learning.}}}
To establish reliable semantic priors, we first construct textual distributions capturing the semantic diversity across classes.
For each class $c$, the base prompt ``a 3D object of \verb|{class}|'' is expanded by an LLM into $M$ paraphrases, producing embeddings $\{\mathbf{z}^{c,1}, \dots, \mathbf{z}^{c,M}\}$ that reflect varied conceptual descriptions of the same category.
The empirical mean and covariance of class $c$ are then computed as:
\begin{align}
\bar{\mathbf{z}}^{c} = \frac{1}{M} \sum_{i=1}^{M} \mathbf{z}^{c,i}, \,
\mathbf{S}^{c} = \sum_{i=1}^{M} (\mathbf{z}^{c,i} - \bar{\mathbf{z}}^{c}) (\mathbf{z}^{c,i} - \bar{\mathbf{z}}^{c})^{\top}.
\end{align}

We model each textual prototype $\boldsymbol{\nu}^{c}$ as a Gaussian variable centered at the empirical mean $\bar{\mathbf{z}}^{c}$, reflecting the uncertainty of language representations across prompt variants:
\begin{align}
p(\boldsymbol{\nu}^{c}) = \mathcal{N}(\boldsymbol{\nu}^{c} \mid \bar{\mathbf{z}}^{c}, \beta^{2} \mathbf{I}),
\end{align}
where $\beta$ controls prior variance.
Given a test feature $\mathbf{x}_t$ at time $t$, its likelihood under class $c$ is:
\begin{align}
p(\mathbf{x}_t \mid \boldsymbol{\nu}^{c}, \mathbf{S}^{c}) = \mathcal{N}(\mathbf{x}_t \mid \boldsymbol{\nu}^{c}, \mathbf{S}^{c}),
\end{align}
where $\mathbf{S}^{c}$ represents the intra-class variability of textual embeddings. 
In practice, a shared covariance $\mathbf{S}$ is used for all classes, which is equivalent to imposing a Dirac prior $p(\mathbf{S}^c)=\delta(\mathbf{S}^c-\mathbf{S})$ that treats the covariance as a fixed parameter.
By combining the prior and likelihood, the posterior over textual parameters is obtained as:
\begin{align}
p(\boldsymbol{\nu}^{c}, \mathbf{S}^{c} \mid \mathbf{x}_t) 
\propto p(\mathbf{x}_t \mid \boldsymbol{\nu}^{c}, \mathbf{S}^{c})\, p(\boldsymbol{\nu}^{c}),
\end{align}
which integrates the semantic evidence from both the textual prior distribution and the incoming visual observation.
The deterministic textual prototype used for inference is then derived via Maximum A Posteriori (MAP) estimation:
\begin{align}\label{eq:nu}
\boldsymbol{\nu}^{c}_{\mathsf{MAP}} =
\left( \beta^{-2} \mathbf{I} + M (\mathbf{S}^{c})^{-1} \right)^{-1}
(\mathbf{S}^{c})^{-1} \bar{\mathbf{z}}^{c}.
\end{align}
% This closed-form update provides a probabilistically grounded textual representation that consolidates semantic diversity into a stable class prior.(see derivation in Section~\ref{appx:derivation}) 
The detailed derivation of Eq.~\eqref{eq:nu} is provided in Section~\ref{appx:derivation} of the Supplementary Material.

% \noindent{\textbf{{Geometric distribution learning.}}
% To capture the evolving geometric structure during test-time, we represent each class $c$ as a Gaussian parameter set
% $\boldsymbol{\Theta}_{t}^{c}=\{\boldsymbol{\mu}_{t}^{c},\boldsymbol{\Sigma}_{t}^{c}\}$,
% which is updated sequentially as new samples arrive.
% At initial time ($t{=}0$), the prior of each class distribution is anchored by its textual prototype $\mathbf{z}^{c}$:
% \begin{equation}
% p(\boldsymbol{\mu}_{0}^{c})
% =\mathcal{N}\big(\boldsymbol{\mu}_{0}^{c}\mid \mathbf{z}^{c},\,\alpha^{2}\mathbf{I}\big),
% \qquad
% \boldsymbol{\Sigma}_{0}^{c}=\boldsymbol{\Sigma}^{c},
% \label{eq:prior_0}
% \end{equation}
% where $\alpha$ controls the prior variance and $\boldsymbol{\Sigma}^{c}$ models the intra-class variability.

\noindent{\textbf{{Geometric distribution learning.}}}
With the textual distributions providing semantic priors, we now model an online geometric distribution for each class during test-time, parameterized as a Gaussian set:
$\boldsymbol{\Theta}_{t}^{c}=\{\boldsymbol{\mu}_{t}^{c},\boldsymbol{\Sigma}_{t}^{c}\}$,
which is updated sequentially as new samples arrive.
At the initial moment ($t{=}0$), the prior of each class distribution is anchored by its textual prototype $\bar{\mathbf{z}}^{c}$:
\begin{equation}\label{eq:geo_prior}
p(\boldsymbol{\mu}_{0}^{c})
=\mathcal{N}\big(\boldsymbol{\mu}_{0}^{c}\mid\bar{\mathbf{z}}^{c},\,\alpha^{2}\mathbf{I}\big),
\quad
\boldsymbol{\Sigma}_{0}^{c}=\mathbf{S}^{c},
% \label{eq:prior_0}
\end{equation}
where $\alpha$ controls the prior variance.
The covariance $\mathbf{S}^{c}$ in Eq.~(6), provides an initial estimate of intra-class variability and serves as a semantic prior for geometric adaptation.

At time $t$, given the new observation $\mathbf{x}_t$, the prior is defined as the previous posterior:
\begin{equation}
p(\boldsymbol{\Theta}_{t}^{c}) = p(\boldsymbol{\Theta}_{t-1}^{c}\mid \mathbf{x}_{t-1}).
\label{eq:prior_t}
\end{equation}
The likelihood of observing $\mathbf{x}_t$ under class $c$ is defined as:
\begin{equation}\label{eq:geo_likelihood}
p(\mathbf{x}_{t}\mid\boldsymbol{\Theta}_{t}^{c})
=\mathcal{N}\big(\mathbf{x}_{t}\mid
\boldsymbol{\mu}_{t}^{c},\,\boldsymbol{\Sigma}_{t}^{c}\big).
% \label{eq:likelihood_t}
\end{equation}
Combining the prior and likelihood, the posterior is recursively updated by Bayes’ rule:
\begin{equation}
\begin{aligned}    
p(\boldsymbol{\Theta}_{t}^{c}\mid \mathbf{x}_{t})
&\propto p(\mathbf{x}_{t}\mid \boldsymbol{\Theta}_{t}^{c})
p(\boldsymbol{\Theta}_{t}^c) \\
&\propto
p(\mathbf{x}_{t}\mid \boldsymbol{\Theta}_{t}^{c})
p(\boldsymbol{\Theta}_{t-1}^{c}\mid \mathbf{x}_{t-1}).
\label{eq:posterior_t}
\end{aligned}
\end{equation}
Under Gaussian assumptions, this recursive update admits a closed-form solution:
\begin{equation}
\begin{aligned}
\boldsymbol{\mu}_t^{c} 
&=\boldsymbol{\Sigma}_t^{c}
\Big((\boldsymbol{\Sigma}^{c})^{-1}\mathbf{x}_t
+(\boldsymbol{\Sigma}_{t-1}^{c})^{-1}\boldsymbol{\mu}_{t-1}^{c}\Big), \\
\boldsymbol{\Sigma}_t^{c}
&=\Big((\boldsymbol{\Sigma}_{t-1}^{c})^{-1}+(\boldsymbol{\Sigma}^{c})^{-1}\Big)^{-1}.
\label{eq:closed_form_update}
\end{aligned}
\end{equation}
The detailed derivation of Eq.~\eqref{eq:closed_form_update} is provided in Section~\ref{appx:derivation} of the Supplemental Material.

\subsection{Multimodal bayesian weighting}
With modality–specific posteriors available, we fuse geometry and text under Bayesian model averaging~\cite{bma}.
Let $\boldsymbol{\Omega}=\{(\boldsymbol{\nu}^{c},\mathbf{S}^{c})\}_{c=1}^{C}$ and $\boldsymbol{\Theta}_t=\{(\boldsymbol{\mu}_t^{c},\boldsymbol{\Sigma}_t^{c})\}_{c=1}^{C}$ denote the
class-wise parameter sets for the textual and geometric modalities, respectively.
The overall posterior for class $c$ at time $t$ is:
\begin{equation}
\label{eq:bma_main}
\begin{aligned}
p(c \mid \mathbf{x}_t)
=&\underbrace{p(c \mid \mathbf{x}_t, \boldsymbol{\Omega}^{c})\,
p(\boldsymbol{\Omega}^{c} \mid \mathbf{x}_t)}_{
\text{Textual posterior predictive}~(\text{predictive}\,\times\,\text{evidence})}\\
+&\underbrace{p(c \mid \mathbf{x}_t, \boldsymbol{\Theta}_t^c)\,
p(\boldsymbol{\Theta}_t^c \mid \mathbf{x}_t)}_{
\text{Geometric posterior predictive}~(\text{predictive}\,\times\,\text{evidence})}
\end{aligned}
\end{equation}
Each term represents a modality-specific posterior predictive,
where $p(\boldsymbol{\Omega}^{c} \mid \mathbf{x}_t)$ and $p(\boldsymbol{\Theta}_t^c \mid \mathbf{x}_t)$
serve as Bayesian weights that automatically balance the two modalities.
Under the textual and geometric distributions derived above, we compute their class-conditional posteriors under Gaussian discriminant analysis (GDA). 
Each modality yields a normalized Gaussian posterior over class $c$ as:
\begin{equation}
\begin{aligned}
p(c \mid \mathbf{x}_t, \boldsymbol{\Omega}^{c})
&=\frac{\mathcal{N}(\mathbf{x}_t\mid\boldsymbol{\nu}^{c}_{\mathsf{MAP}},\mathbf{S}^{c})}
{\sum_{c'}\mathcal{N}(\mathbf{x}_t\mid\boldsymbol{\nu}^{c'}_{\mathsf{MAP}},\mathbf{S}^{c'})},\\
p(c \mid \mathbf{x}_t, \boldsymbol{\Theta}_t^c)
&=\frac{\mathcal{N}(\mathbf{x}_t\mid\boldsymbol{\mu}_t^{c},\boldsymbol{\Sigma}_t^{c})}
{\sum_{c'}\mathcal{N}(\mathbf{x}_t\mid\boldsymbol{\mu}_t^{c'},\boldsymbol{\Sigma}_t^{c'})}.
\end{aligned}
\label{eq:gda_post}
\end{equation}
Substituting Eq.~\eqref{eq:gda_post} into the Eq.~\eqref{eq:bma_main} yields the final multimodal posterior $p(c\mid\mathbf{x}_t)$.

\section{Experiments}
\begin{table*}[t]
% \footnotesize
% \tabstyle{5pt}
\caption{{Recognition accuracy comparison on ModelNet-C with 7 corruption types.} Each clean point cloud contains 1024 points, and the corruption severity level is set to 2. The last column reports the average accuracy over all types. 
The best results are highlighted in \textbf{bold}, and the second best are \underline{underlined}. 
% This setting applies to the following tables unless otherwise specified.
}\label{tab:modelnet_c_robustness}
 \centering
   \begin{adjustbox}{width=1.\linewidth}
   % \resizebox{1.\linewidth}{!}{
   \begin{tabular}{l c c c c c c c c c}
      \toprule
      \multirow{2}{*}{Method} & \textbf{Clean Data} & \multicolumn{7}{c}{\textbf{Corruption Type}} & \multirow{2}{*}{\textbf{Avg.}} \\\cmidrule{3-9}
            & ModelNet & Add Global & Add Local & Drop Global & Drop Local & Rotate & Scale & Jitter & \\
      \midrule
ULIP~\cite{xue23ulip} & 56.16 & 33.55 & 43.92 & 54.70 & 50.89 & 55.27 & 50.20 & 44.08 & 48.60 \\
\ + Point-Cache (Global) & 62.12 & 45.79 & \underline{47.98} & 56.85 & 53.89 & 60.25 & 54.34 & 48.91 & 53.77 \\ 
\ + Point-Cache (Hierarchical) & \underline{64.22} & \underline{46.15} & 47.85 & \underline{59.16} & \underline{56.00} & \underline{61.47} & \underline{55.35} & \underline{49.92} & \underline{55.02} \\ 
\ + \textbf{\Ours} & \textbf{66.04} & \textbf{54.82} & \textbf{53.93} & \textbf{63.09} & \textbf{60.13} & \textbf{63.82} & \textbf{60.49} & \textbf{53.04} & \textbf{59.42} \\ 
\midrule
ULIP-2~\cite{xue24ulip2} & 71.23 & 65.15 & 54.62 & 68.76 & 57.98 & 70.30 & 67.10 & 21.76 & 59.61 \\ 
\ + Point-Cache (Global) & 73.95 & 67.02 & 59.32 & 71.35 & 61.59 & 72.37 & 68.40 & 28.20 & 62.78 \\
\ + Point-Cache (Hierarchical) & \underline{74.53} & \underline{68.11} & \underline{61.26} & \textbf{73.22} & \underline{63.65} & \underline{73.34} & \underline{70.42} & \textbf{29.50} & \underline{64.25} \\ 
\ + \textbf{\Ours} & \textbf{76.30} & \textbf{69.04} & \textbf{64.38} & \underline{72.57} & \textbf{64.18} & \textbf{74.55} & \textbf{71.64} & \underline{29.01} & \textbf{65.21} \\ 
\midrule
O-Shape~\cite{liu23openshape} & \underline{84.56} & 71.64 & 67.79 & 81.56 & 73.58 & 82.01 & 78.48 & 59.36 & 74.87 \\
\ + Point-Cache (Global) & 84.52 & 74.72 & 72.77 & 82.41 & 75.12 & 83.18 & 78.93 & 67.91 & 77.45 \\
\ + Point-Cache (Hierarchical) & 84.04 & \underline{74.84} & \underline{73.70} & \underline{82.21} & \underline{76.26} & \underline{82.66} & 78.12 & \underline{68.35} & \underline{77.52} \\ 
\ + \textbf{\Ours} & \textbf{85.49} & \textbf{75.36} & \textbf{74.39} & \textbf{83.14} & \textbf{77.07} & \textbf{84.08} & \textbf{79.86} & \textbf{69.45} & \textbf{78.61} \\ 
\midrule
Uni3D~\cite{zhou24uni3d} & 81.81 & 72.45 & 56.36 & 68.15 & 67.18 & 79.94 & 75.36 & 56.24 & 69.69 \\ %
\ + Point-Cache (Global) & 83.14 & 76.13 & 66.49 & 71.43 & 69.81 & 81.52 & 75.85 & 61.43 & 73.20 \\
\ + Point-Cache (Hierarchical) & \underline{83.87} & \underline{77.51} & \underline{71.15} & \underline{72.16} & \underline{70.75} & \underline{81.77} & \underline{77.31} & \underline{62.52} & \underline{74.63} \\ 
\ + \textbf{\Ours} & \textbf{85.17} & \textbf{77.59} & \textbf{73.30} & \textbf{74.96} & \textbf{71.88} & \textbf{83.75} & \textbf{79.98} & \textbf{65.84} & \textbf{76.56} \\ 
\bottomrule
   \end{tabular}
   \end{adjustbox}
    % }
\end{table*}

\begin{table*}[t]
   % \footnotesize
   \tabstyle{6pt}
   \centering
\caption{{Recognition accuracy comparison on multiple benchmarks.} 
   S-PB\_RS\_T50 denotes the hardest split of ScanObjectNN. 
   O-LVIS and Omni3D refer to Objaverse-LVIS and OmniObject3D, respectively. 
   The number under each dataset indicates the number of points per object (pts). 
   In Omni3D, each object may contain a variable number of points.  }
   \label{tab:multi_dataset_generalization}
   % \begin{adjustbox}{width=0.9\linewidth}
   \begin{tabular}{l c c c c c c c c }
      \toprule
      \multirow{2}{*}{Method} & ModelNet40 & S-PB\_RS\_T50 & O-LVIS & \multicolumn{3}{c}{Omni3D} & \multirow{2}{*}{\textbf{Avg.}} \\\cmidrule{5-7} %
      & (10000 pts) & (2048 pts) & (10000 pts) & (1024 pts & 4096 pts & 16384 pts) \\
      \midrule
ULIP~\cite{xue23ulip} & 58.75 & 46.44 & 6.24 & 8.39 & 7.75 & 7.28 & 22.48 \\
\ + Point-Cache (Global) & 61.22 & 50.21 & \underline{7.02} & 10.00 & 9.36 & 8.43 & 24.37 \\
\ + Point-Cache (Hierarchical) & \underline{62.93} & \underline{51.80} & \underline{7.02} & \underline{10.47} & \underline{9.75} & \underline{8.90} & \underline{25.15} \\
\ + \textbf{\Ours} & \textbf{67.13} & \textbf{53.67} & \textbf{7.79} & \textbf{11.27} & \textbf{10.68} & \textbf{9.47} & \textbf{26.67}\\
\midrule
ULIP-2~\cite{xue24ulip2} & 72.97 & 47.13 & 30.26 & 26.36 & 29.20 & 26.58 & 38.75 \\
\ + Point-Cache (Global) & 74.51 & 51.70 & \underline{32.65} & 28.51 & 31.10 & 28.53 & 41.17 \\
\ + Point-Cache (Hierarchical) & \underline{75.53} & \underline{54.98} & 32.36 & \textbf{29.37} & \underline{31.24} & \underline{29.44} & \underline{42.15} \\
\ + \textbf{\Ours} & \textbf{76.78} & \textbf{56.47} & \textbf{32.76} & \underline{28.98} & \textbf{31.76} & \textbf{31.12} & \textbf{42.98}\\
\midrule
      O-Shape~\cite{liu23openshape} & 84.52 & 54.60 & \underline{46.78} & 33.21 & 33.52 & 33.37 & 47.67\\ %
      \ + Point-Cache (Global) & 85.70 & \underline{57.13} & \textbf{47.03} & \underline{36.92} & 37.61 & \underline{37.43} & \underline{50.30} \\
      \ + Point-Cache (Hierarchical) & \textbf{85.90} & 56.61  & 45.63 & 36.87 & \underline{38.02} & 37.39 & 50.07\\
      \ + \textbf{\Ours} & \underline{85.74} & \textbf{66.12} & 43.93 & \textbf{37.77} & \textbf{38.03} & \textbf{38.38} & \textbf{51.66}\\
\midrule
      Uni3D~\cite{zhou24uni3d} & 88.41 & 65.19 & \textbf{55.42} & 31.52 & 41.98 & 41.86 & 54.09 \\ %
      \ + Point-Cache (Global) & 88.86 & \underline{68.51} & 53.36 & 34.97 & 45.13 & 45.19 & 56.00 \\
      \ + Point-Cache (Hierarchical) & \underline{89.18} & 68.24 & \underline{55.19} & \underline{35.82} & \underline{45.60} & \underline{45.89} & \underline{56.65} \\
      \ + \textbf{\Ours} & \textbf{90.48} & \textbf{73.04} & 53.63 & \textbf{36.54} & \textbf{45.97} & \textbf{46.68} &\textbf{57.72} \\
      
      \bottomrule
   \end{tabular}
   % \end{adjustbox}
\end{table*}
\begin{table*}[ht]
    \centering
    % \footnotesize
    \tabstyle{8pt}
    \caption{{Recognition accuracy comparison on Sim-to-Real}. Two evaluation settings are considered:  
    MN\_11 $\rightarrow$ SONN\_11 and SN\_9 $\rightarrow$ SONN\_9. 
    The dataset on the left side of $\rightarrow$ stands for simulated data, while the dataset on the right side indicates real-world data.  
    11 classes are shared between MN\_11 and SONN\_11, while 9 classes are common between SN\_9 and SONN\_9. 
    In the experiments, each point cloud is represented by 2,048 points. 
    MN: ModelNet, SN: ShapeNet.}
% \begin{adjustbox}{width=1\linewidth}
\begin{tabular}{l c c c c c c c c}
    \toprule
    \multirow{2}{*}{Method} & \multicolumn{3}{c}{MN\_11 $\rightarrow$ SONN\_11} & & \multicolumn{3}{c}{SN\_9 $\rightarrow$ SONN\_9} & \multirow{2}{*}{\textbf{Avg.}} \\\cmidrule{2-4}\cmidrule{6-8}
          & OBJ & OBJ\_BG & PB\_T50\_RS & & OBJ & OBJ\_BG & PB\_T50\_RS\\
    \midrule

ULIP~\cite{xue23ulip} & 57.05 & 50.32 & 32.60 & & 61.00 & 61.00 & 44.38 & 51.06 \\
    \ + Point-Cache (Global) & 62.32 & 52.63 & 34.97 & & \underline{65.50} & 62.50 & 47.36 & 54.21 \\
   \ + Point-Cache (Hierarchical) & \underline{64.42} & \underline{56.63} & \underline{35.77} & & \textbf{67.25} & \underline{64.50} & \underline{47.61} & \underline{56.03} \\
   \ + \textbf{\Ours} & \textbf{65.41} & \textbf{60.42} & \textbf{45.34} & & \textbf{67.25} & \textbf{68.50} & \textbf{55.36} & \textbf{60.38} \\
    \midrule
ULIP-2~\cite{xue24ulip2} & 50.94 & 52.42 & 39.12 & & 51.50 & 59.25 & 46.35 & 49.93 \\ 
+ Point-Cache (Global) & 55.10 & 58.52 & 47.38 & & 56.75 & \underline{65.00} & 50.68 & 55.57 \\ 
+ Point-Cache (Hierarchical) & \underline{57.26} & \underline{58.95} & \underline{47.71} & & \underline{58.00} & \textbf{70.25} & \underline{52.70} & \underline{57.48} \\
+ \textbf{\Ours} & \textbf{57.89} & \textbf{58.97} & \textbf{50.25} & & \textbf{61.25} & 64.25 & \textbf{56.22} & \textbf{58.14} \\
\midrule

O-Shape~\cite{liu23openshape} & 59.78 & 62.53 & 45.51 &  & 64.00 & 70.25 & 53.55 & 59.27 \\
+ Point-Cache (Global) & 65.07 & 68.67 & 46.23 & & 71.00 & \underline{71.50} & \underline{55.67} & \underline{63.02} \\
+ Point-Cache (Hierarchical) & \underline{66.11} & \underline{69.68} & \underline{47.50} & & \underline{71.50} & 71.00 & 56.57 & 63.78 \\
+ \textbf{\Ours} & \textbf{69.05} & \textbf{71.76} & \textbf{56.94} & & \textbf{71.75} & \textbf{74.50} & \textbf{64.00} & \textbf{68.00} \\
\midrule

Uni3D~\cite{zhou24uni3d} & 72.63 & 74.53 & 55.76 &  & 67.50 & 68.50 & 57.98 & 66.18 \\
+ Point-Cache (Global) & \textbf{76.21} & \textbf{77.26} & \underline{59.10} & & \underline{74.50} & \underline{76.50} & \underline{62.47} & \textbf{71.01} \\
+ Point-Cache (Hierarchical) & 74.11 & 76.00 & 57.92 & & \textbf{77.50} & \textbf{78.00} & 58.89 & 69.07 \\
+ \textbf{\Ours} & \underline{74.31} & \underline{77.05} & \textbf{62.48} & & 72.50 & 76.00 & \textbf{63.32} & \underline{70.94} \\
\bottomrule

\end{tabular}
% \end{adjustbox}
    \label{tab:xset_sim2real_generalization}
\end{table*}
\subsection{Experimental settings}
\label{subsec:experiment_settings}

\noindent\textbf{Datasets.}
To evaluate the \emph{robustness} of point cloud recognition, we adopt four public datasets covering diverse corruption types. 
Specifically, we use ModelNet-C~\cite{ModelNet-C} and three corrupted variants of ScanObjectNN-C~\cite{ScanObjectNN-C}.
ModelNet-C defines seven atomic corruptions, including \emph{global outliers}, \emph{local outliers}, \emph{global structure dropping}, \emph{local part dropping}, \emph{rotation}, \emph{scaling}, and \emph{jittering}, from which other corruption types can be derived.
Following~\cite{ModelNet-C}, we apply these atomic corruptions to the three variants of ScanObjectNN to construct their corrupted versions.
To assess \emph{generalization} on unseen data, we further test our method on four challenging benchmarks: OmniObject3D~\cite{OmniObject3D} (216 classes), Objaverse-LVIS~\cite{Objaverse} (1,156 classes), the hardest variant of ScanObjectNN, and ModelNet40~\cite{3D_ShapeNets}.
Recognition accuracy (\%) is reported as the main metric.

\noindent\textbf{Models.}
We evaluate our approach on four representative multimodal 3D foundation models: ULIP~\cite{xue23ulip}, ULIP-2~\cite{xue24ulip2}, OpenShape~\cite{liu23openshape}, and Uni3D~\cite{zhou24uni3d}.
All models are initialized with publicly released pre-trained weights and remain \emph{frozen} during evaluation.
For fair comparison, each model operates on point clouds uniformly sampled to 1,024 points and normalized within the unit sphere.
All experiments are implemented under a unified evaluation framework, ensuring consistent preprocessing and input configurations
Further implementation details can be found in Section~\ref{appx:implement} of the Supplementary Material.

\subsection{Main results}
\label{subsec:comparison_experiments_robustness}

\noindent\textbf{Robustness.}
We evaluate the test-time robustness of our method on ModelNet-C~\cite{ModelNet-C}, which includes seven corruption types in addition to the clean set.
As shown in Table~\ref{tab:modelnet_c_robustness}, our training-free \texttt{BayesMM} significantly enhances the robustness of all four large multimodal 3D backbones.
When averaged over the clean and corrupted settings, \texttt{BayesMM} improves the performance of ULIP~\cite{xue23ulip} from 48.60\% to 59.42\% (+10.82), ULIP-2~\cite{xue24ulip2} from 59.61\% to 65.21\% (+5.60), OpenShape~\cite{liu23openshape} from 74.87\% to 78.61\% (+3.74), and Uni3D~\cite{zhou24uni3d} from 69.69\% to 76.56\% (+6.87).
Compared with Point-Cache baselines, \texttt{BayesMM} consistently achieves the highest average accuracy across all corruption types, demonstrating stronger robustness against structural perturbations, outliers, rotations, scale variations, and geometric distortions.

Notably, our $\texttt{\Ours}$ not only mitigates degradation on corrupted data but also improves recognition on clean inputs.
For instance, ULIP and ULIP-2 gain +9.88\% and +5.07\% absolute improvements on the clean ModelNet dataset, respectively.
While OpenShape retains the highest clean accuracy among all models, our cache mechanism yields larger robustness margins under corruption, outperforming its original model by +5.12\% on average.
Similar consistent gains are observed across all corruption types and across different 3D backbones, demonstrating that our hierarchical caching scheme generalizes well across architectures without any retraining or fine-tuning.
Further results on ScanObjectNN-C variants are provided in Table~\ref{tab:s_obj_only_c_robustness}, \ref{tab:s_obj_bg_c_robustness}, and \ref{tab:s_hardest_c_robustness} in the Supplementary.
\label{subsec:comparison_experiments_generalization}

\noindent\textbf{Generalization.}
To further evaluate the generalization ability of our method under domain and distribution shifts, 
we test it across four benchmarks: ModelNet40~\cite{3D_ShapeNets}, 
the hardest split of ScanObjectNN (S-PB\_RS\_T50)~\cite{ScanObjectNN-C}, 
Objaverse-LVIS (O-LVIS)~\cite{Objaverse}, and OmniObject3D~\cite{OmniObject3D}, 
as summarized in Table~\ref{tab:multi_dataset_generalization}. 
These datasets span diverse object categories, point densities, and real-world variations, 
providing a comprehensive evaluation of model robustness and transferability. 
\texttt{BayesMM} consistently improves recognition accuracy across all datasets and architectures 
without any retraining or fine-tuning. 
For ULIP~\cite{xue23ulip}, the average accuracy rises from 22.48\% to 26.67\% (+4.19), 
with similar gains for ULIP-2~\cite{xue24ulip2} (+4.23), OpenShape~\cite{liu23openshape}(+3.99) and Uni3D~\cite{zhou24uni3d} (+3.63).  On the Omni3D dataset, \texttt{BayesMM} exhibits strong generalization across varying point densities (1024, 4096, 16384), achieving substantial performance gains under all settings.
Even when the baselines already achieve strong performance 
(\eg, Uni3D on ModelNet40 and the hardest split of ScanObjectNN), 
\texttt{BayesMM} continues to deliver steady improvements.

\noindent\textbf{Generalization from simulated to real data.}
Following Sim-to-Real~\cite{sim2real}, we further evaluate our method under the cross-domain setting from simulated to real-world data, covering two scenarios: MN\_11~$\rightarrow$~SONN\_11 and SN\_9~$\rightarrow$~SONN\_9.  
As shown in Table~\ref{tab:xset_sim2real_generalization}, \texttt{BayesMM} consistently improves recognition performance across all foundation backbones.  
Compared with Point-Cache~\cite{Point-Cache}, \texttt{BayesMM} ranks at or near the top in almost all metrics, demonstrating stronger generalization of pre-trained 3D models without any addiandtional training.  
On ULIP~\cite{xue23ulip}, our method achieves an average accuracy of 60.38\%, outperforming the hierarchical Point-Cache by +4.35\%.  
Consistent gains are also observed on ULIP-2~\cite{xue24ulip2} (+0.66\%), OpenShape~\cite{liu23openshape} (+4.22\%), and Uni3D~\cite{zhou24uni3d} (+1.87\%), indicating that the proposed distribution learning strategy generalizes robustly across architectures of varying capacity.  
Notably, \texttt{BayesMM} achieves clear gains on the PB\_T50\_RS subsets, reflecting its enhanced robustness to real-world noise, background clutter, and geometric perturbations.

\subsection{Ablation study}
\label{subsec:ablation_study}

%4.3Memory and Usage Thtoughput这个标题不要，全部放在Ablation study.
%二级标题只有第一个词首字母大写
%Table 4横纵换一下
%再加一个主实验的表，内容似乎不够, 似乎要放两个
%所有的表格新建.text文件，放在.text文件里
\begin{table}[!t]
\caption{Ablation study on the components of \Ours. S-OBJ denotes the OBJ\_ONLY split of ScanObjectNN.}
\centering
\tabstyle{2pt}  
\begin{tabular}{lccccc}
\toprule
& \multirow{2}{*}{\makecell{Geometric\\Distribution}} & \multirow{2}{*}{\makecell{Texutal\\Distribution}} & \multirow{2}{*}{\makecell{Bayes\\Weighting}} & \multicolumn{2}{c}{ULIP2~\cite{xue24ulip2}} \\  \cmidrule(lr){5-6}
 &  &  &  & S-OBJ  & Omni3D  \\ 
\midrule
(1) & \ding{55} & \ding{55}  & \ding{55}  & 42.00 & 26.58  \\ 
\midrule
(2) & \ding{52} & \ding{55} & \ding{55}  & 46.47   & 26.63   \\ 
\midrule
(3) & \textbf{\ding{55}} & \textbf{\ding{52}} & \ding{55}  & \underline{52.50}  & \underline{30.79}\\ \midrule
(4) & \textbf{\ding{52}} & \textbf{\ding{52}} & \ding{52}  & \textbf{53.02}  & \textbf{31.12}\\
\bottomrule
\end{tabular}
\label{tab:diff_modules}
\end{table}

\begin{figure}[t]
  \centering
  \begin{subfigure}{0.475\linewidth}
    \centering
    \includegraphics[width=\linewidth]{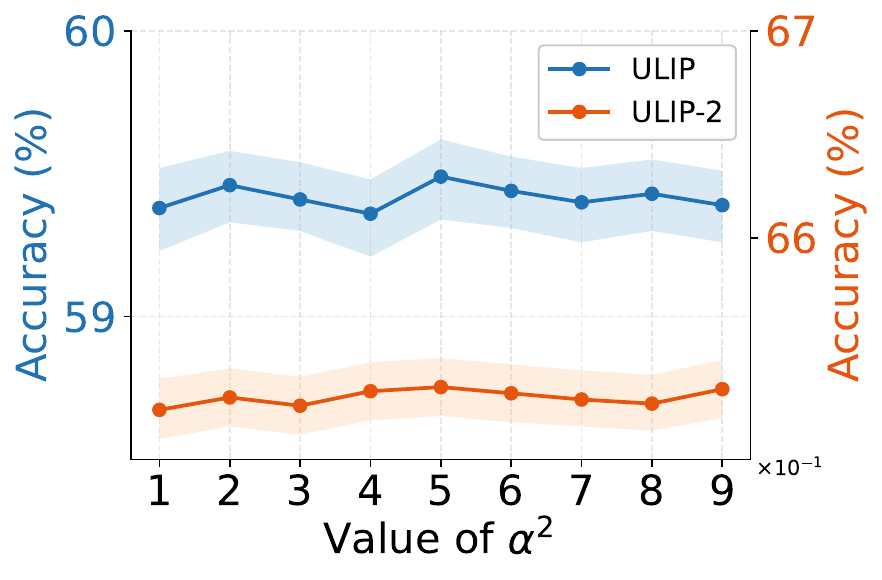}
    \caption{Effect of $\alpha^2$ on geometric distribution learning}
    \label{fig:alpha}
  \end{subfigure}
  \hfill
  \begin{subfigure}{0.475\linewidth}
    \centering
    \includegraphics[width=\linewidth]{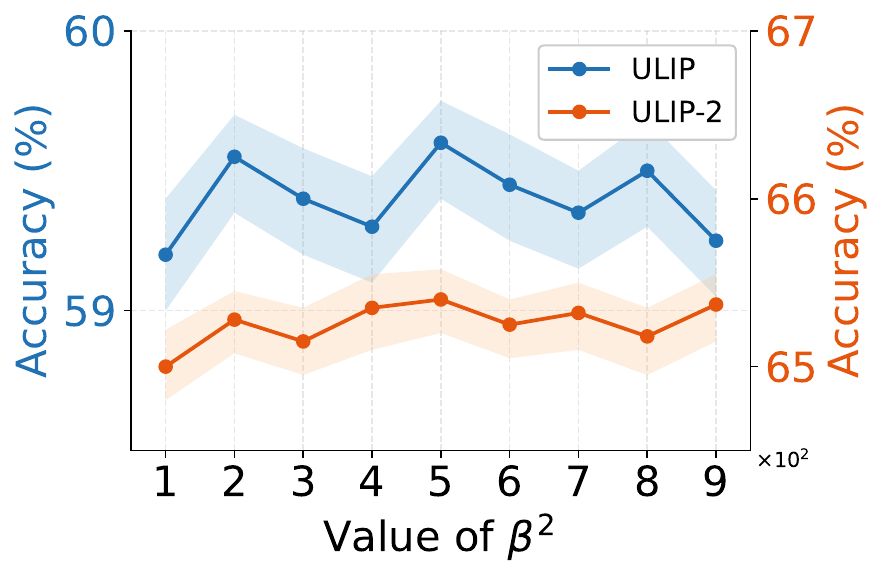}
    \caption{Effect of $\beta^2$ on textual distribution learning}
    \label{fig:beta}
  \end{subfigure}
  \caption{
  Hyperparameter sensitivity of \texttt{BayesMM} with respect to the geometric prior $\alpha^2$ 
  and textual prior $\beta^2$, evaluated on ULIP and ULIP-2 backbones.}
  \label{fig:hyperparam}
\end{figure}

\noindent\textbf{Ablation on model components.}
To investigate the contribution of each component within \texttt{BayesMM}, 
we perform an ablation study on the ULIP2~\cite{xue23ulip}backbone, evaluated on the ScanObjectNN (OBJ\_ONLY split, denoted as S-OBJ) and Omni3D benchmarks, as summarized in Table~\ref{tab:diff_modules}.  
The baseline using only fixed text prototypes performs poorly due to the lack of adaptive refinement.  
Introducing geometric distribution learning significantly improves accuracy, as the recursive update in Eq.~\eqref{eq:closed_form_update} enables online prototype refinement.  
Adding textual distribution learning further boosts performance by modeling intra-class semantic variability with LLM-generated paraphrases.  
Combining both modalities under the Bayesian model averaging principle (Eq.~\eqref{eq:bma_main}) achieves the best results, showing that geometric and textual distributions are complementary—geometric modeling enhances adaptability, textual modeling enriches semantics, and Bayesian weighting integrates them coherently within a unified probabilistic framework.

\noindent{\textbf{Sensitivity analysis of $\alpha^2$ and $\beta^2$.}} 
To examine the sensitivity of \texttt{BayesMM} to hyperparameter choices, 
$\alpha^2$ and $\beta^2$ are varied to control the weighting strength in Bayesian fusion and the variance scale in multimodal distribution learning, respectively.  
As shown in Figure~\ref{fig:hyperparam}(a), varying $\alpha^2$ has minimal impact on accuracy. 
Both ULIP~\cite{xue23ulip} and ULIP-2~\cite{xue24ulip2} remain stable, indicating that the geometric prior is insensitive to moderate changes in weighting strength and that Bayesian fusion effectively balances geometric and textual cues.  
Figure~\ref{fig:hyperparam}(b) shows a similar trend for $\beta^2$, where accuracy remains nearly constant across different variance scales. 
ULIP~\cite{xue23ulip} stays around 59.4\%, while ULIP-2~\cite{xue24ulip2} remains near 65.2\%, showing only minor fluctuations. 
These results demonstrate that \texttt{BayesMM} is robust to hyperparameter variations and maintains consistent performance across models and datasets without fine-tuning.

\begin{figure}[t]
  \centering
  \begin{subfigure}{0.45\linewidth}
  \centering
    \includegraphics[width=1.02\linewidth]{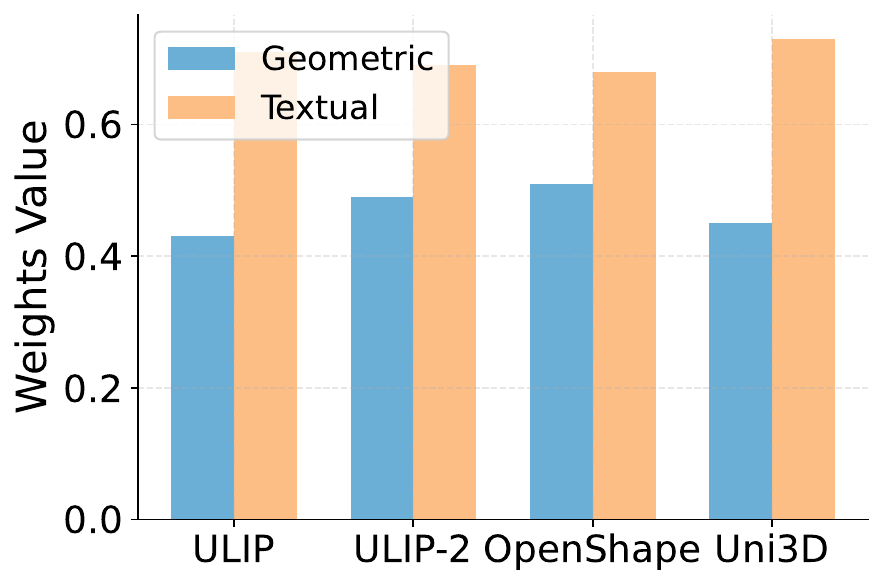}
    \caption{Weight comparison on clean data}
    \label{fig:img_aug}
  \end{subfigure}
\hfill 
 \begin{subfigure}{0.45\linewidth}
    \centering
    \includegraphics[width=1.02\linewidth]{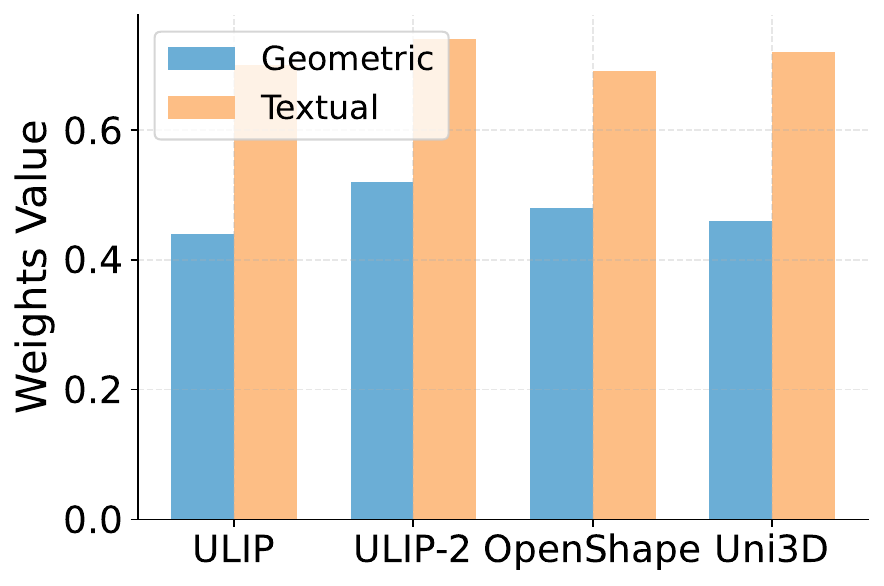}
    \caption{Weight comparison on corruption data}
    \label{fig:txt_aug}
  \end{subfigure}
  \caption{ Analysis of geometric and textual weights in \texttt{BayesMM} across different backbones on the ModelNet-C dataset.} 
  \label{fig:bayes_weight}
\end{figure}
\noindent{\textbf{Analysis of bayesian weighting.}} To better understand the role of modality balancing in \texttt{BayesMM},
we examine the learned Bayesian weights across different models on the ModelNet-C dataset.
As shown in Figure~\ref{fig:bayes_weight}, the textual modality consistently receives higher weights than the geometric one,
reflecting its stronger stability and semantic generalization under distribution shifts.
This weighting pattern remains consistent across both clean and corrupted inputs,
demonstrating that \texttt{BayesMM} adaptively calibrates cross-modal contributions
and effectively suppresses noise sensitivity in the geometric stream, leading to more reliable multimodal inference.

\subsection{Memory Usage and Throughput}
\begin{table}[ht]
   \footnotesize
   \centering
   \caption{
   Memory usage (MB) comparison across different datasets. 
Numbers below each dataset name indicate the number of classes. 
Point-Cache denotes the hierarchical variant.
}
   \begin{tabular}{l c c c c}
      \toprule
      \multirow{2}{*}{Method} & ModelNet-C & Omni3D & O-LVIS  & \#Params\\
      & (40) & (216) & (1156) &(M)\\
      \midrule
      ULIP~\cite{xue23ulip} & \textbf{1,556} & \textbf{1,558} & \textbf{1,556} &85.7 \\ 
      \ + Point-Cache & \textbf{1,556} & \textbf{1,558} & 1,566 &85.7 \\
      \ + \textbf{\Ours} & \underline{1,560} & \underline{1,560} & \underline{1,562} &85.7 \\
      \midrule
      Uni3D~\cite{zhou24uni3d} & \textbf{5,062} & \textbf{5,062} & \textbf{5,062} &1016.5 \\ 
      \ + Point-Cache & \underline{5,064} & \underline{5,068} & 5,090 &1016.5\\
      \ + \textbf{\Ours} & 5,076 & 5,077 & \underline{5,080} &1016.5 \\
      \bottomrule
   \end{tabular}
   \label{tab:memory_comparison_uni3d}
\end{table}
\noindent\textbf{Memory.}  
We evaluate the memory consumption of our method compared with Point-Cache and the corresponding baselines ULIP~\cite{xue23ulip} and Uni3D~\cite{zhou24uni3d}, as summarized in Table~\ref{tab:memory_comparison_uni3d}.  
Although our approach introduces a slightly higher memory footprint at ModelNet-C~\cite{ModelNet-C}, its growth with respect to the number of categories is significantly slower.  
For example, when scaling from ModelNet-C (40 classes)~\cite{ModelNet-C} to O-LVIS (1,156 classes)~\cite{Objaverse}, the total memory usage of Uni3D increases by nearly \( +18\,\text{MB} \) under Point-Cache, whereas our hierarchical cache only adds about \( +4\,\text{MB} \).  
This trend demonstrates that our method effectively amortizes the class-wise parameter overhead by sharing covariance structures across categories.  
The result is a nearly constant per-class memory cost even in large-scale scenarios.  
Furthermore, the total memory remains dominated by the heavy backbone parameters of Uni3D (\(\sim\!1{,}016.5\,\text{M}\)), making the additional cost of our method negligible.

\noindent\textbf{Throughput.}
We further compare inference throughput across different models.
As shown in Table~\ref{tab:throughput_comparison}, \texttt{BayesMM} introduces only marginal overhead relative to Point-Cache and zero-shot baselines.
Although additional refinement is performed, these operations are parallelizable on GPUs, yielding minimal runtime impact.
Overall, our method preserves over 97\% of the zero-shot inference speed while providing substantial gains in accuracy and robustness.

% \noindent\textbf{Throughput.}
% We further compare the inference throughput (measured as the number of samples processed per second) of different models.
% As shown in Table~\ref{tab:throughput_comparison}, our method introduces only a marginal overhead compared to Point-Cache and the zero-shot baselines.
% Although our method performs additional refinement, these operations are highly parallelizable on GPUs, making the actual runtime overhead even smaller than what the throughput numbers suggest.
% Consequently, the throughput remains close to that of the baseline across all models.
% In particular, our approach maintains over 97\% of the zero-shot inference speed while achieving substantial improvements in accuracy and robustness, thus demonstrating a favorable balance between efficiency and adaptability.
\begin{table}[!t]
   % \footnotesize
   \tabstyle{4pt}  
   \centering
\caption{%
Inference throughput (samples per second) on the ModelNet40. 
All experiments are conducted with a batch size of 1 on an RTX 3090 GPU. 
Point-Cache denotes the hierarchical variant.
   }
   \begin{tabular}{l c c c c}
      \toprule
      {Method} & ULIP & ULIP2 & OpenShape & Uni3D\\
      \midrule
       Vanilla  & \textbf{11.35} & \textbf{12.82}& \textbf{8.68} & \textbf{7.77} \\ 
      \ + Point-Cache~\cite{Point-Cache}  & \underline{11.27}& \underline{12.75} & \underline{8.60} & \underline{7.69} \\
      \ + \textbf{\Ours}  & 10.99 & 12.45 & 8.29 & 7.45 \\
      \bottomrule
   \end{tabular}
   \label{tab:throughput_comparison}
\end{table}

\subsection{Visualization}
\begin{figure}[t]
  \centering
  \begin{subfigure}{0.49\linewidth}
  \centering
    \includegraphics[width=1\linewidth]{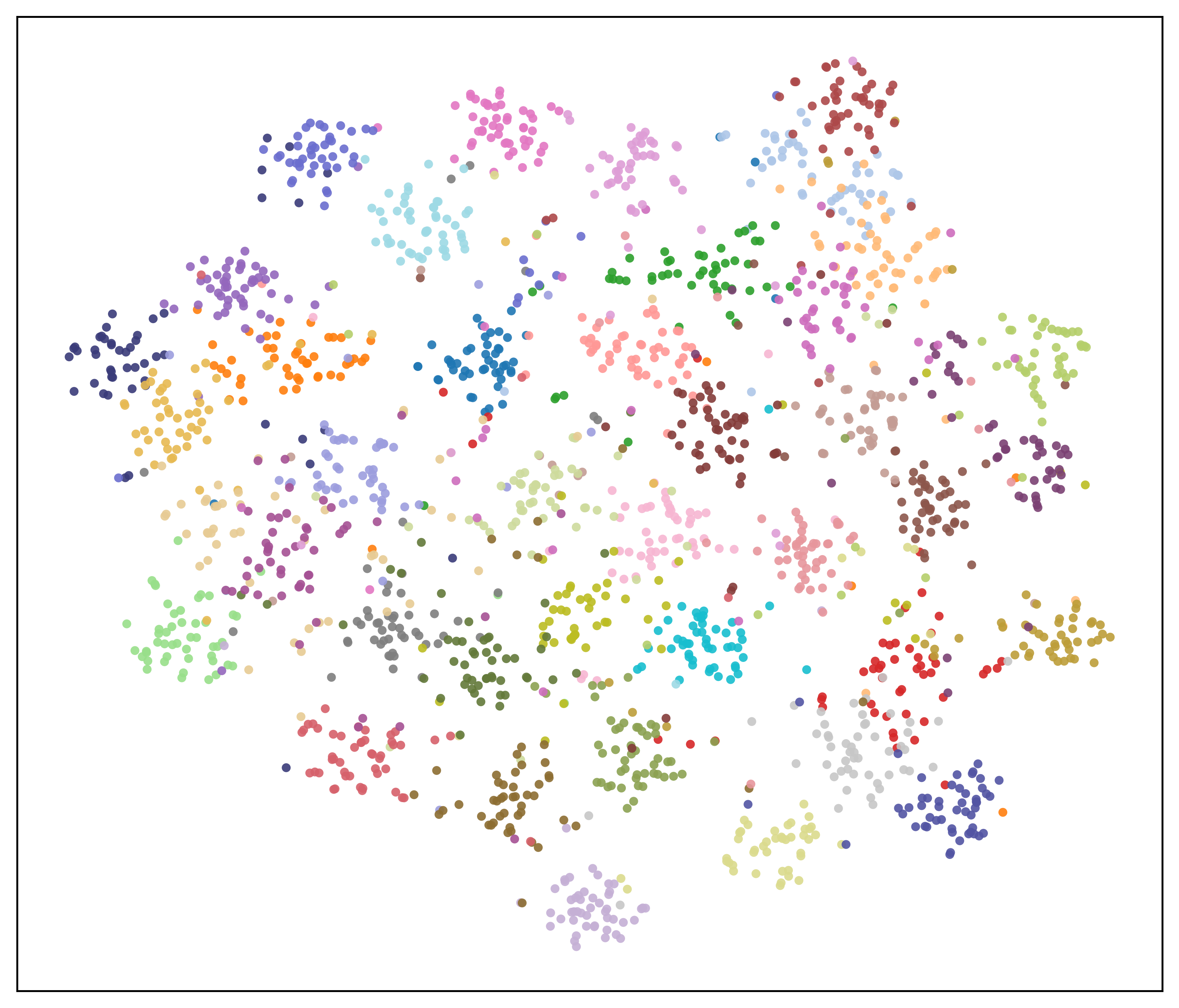}
    \caption{Cache-based updating.}
    \label{fig:img_aug}
  \end{subfigure}
 \begin{subfigure}{0.49\linewidth}
    \centering
    \includegraphics[width=1\linewidth]{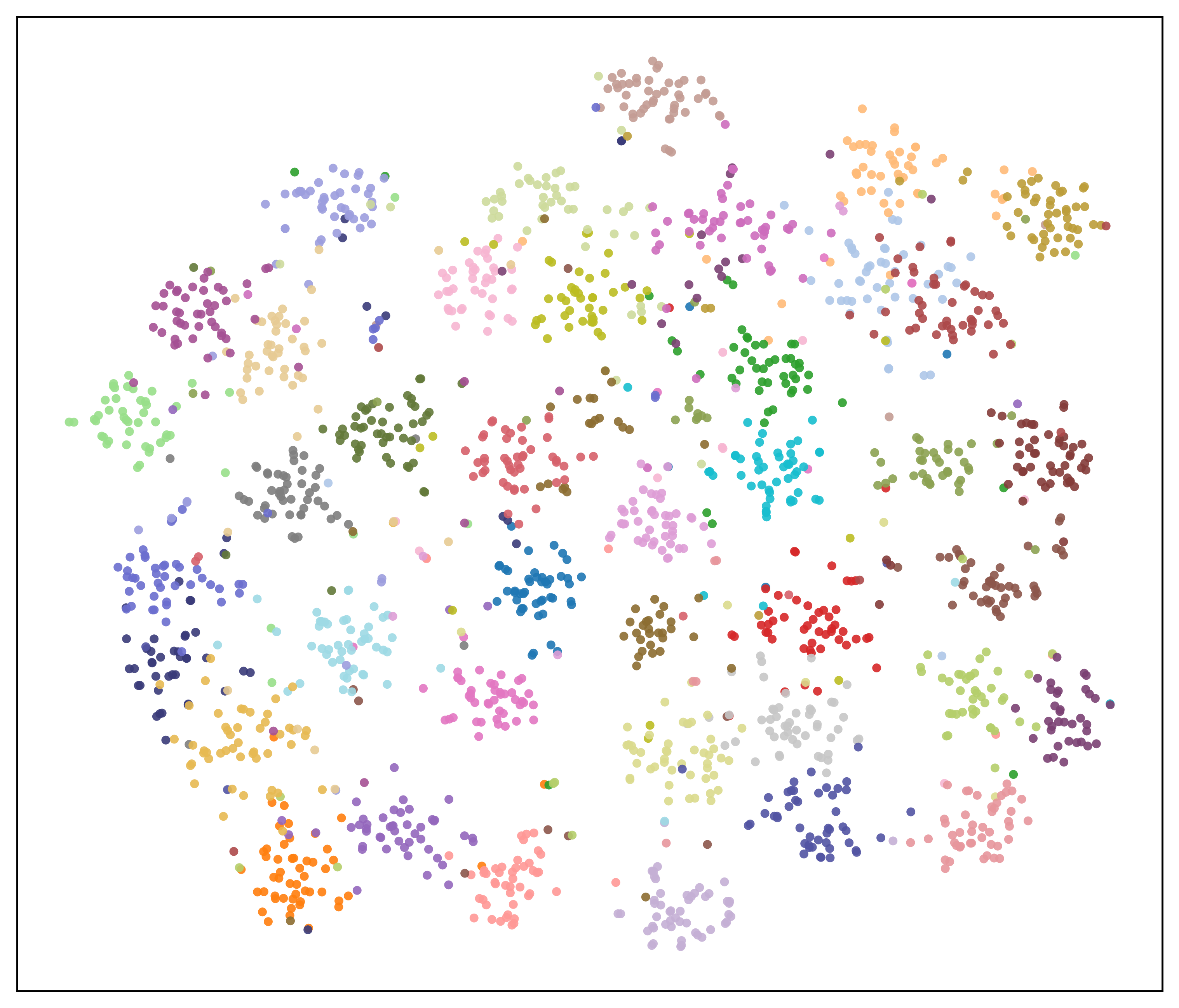}
    \caption{Geometric distribution learning.}
    \label{fig:txt_aug}
    \end{subfigure}
  \label{fig:aug}
   \begin{subfigure}{0.49\linewidth}
    \centering
    \includegraphics[width=1\linewidth]{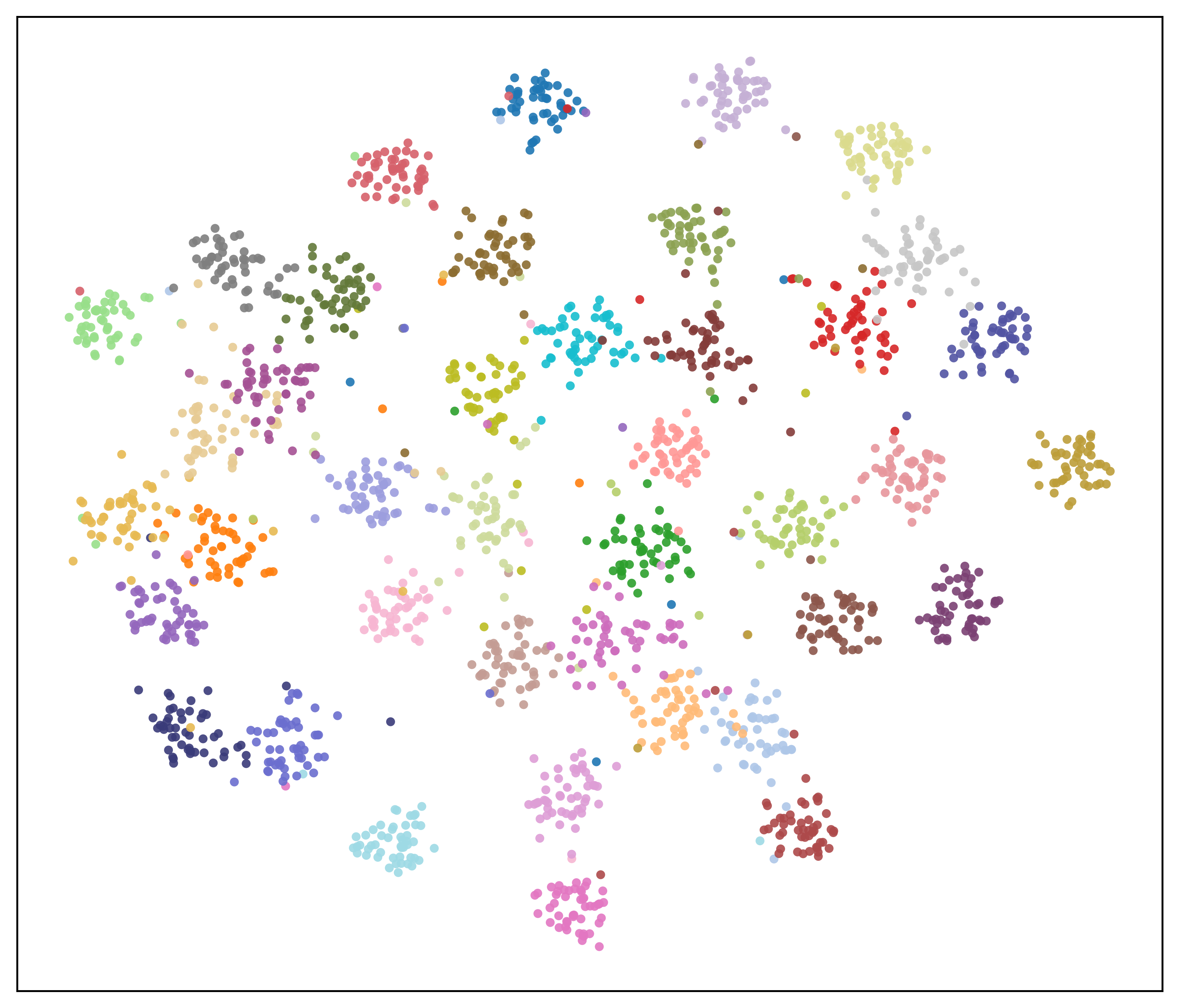}
    \caption{Textual distribution learning.}
    \label{fig:txt_aug}
  \end{subfigure}
  \label{fig:aug}
   \begin{subfigure}{0.49\linewidth}
    \centering
    \includegraphics[width=1\linewidth]{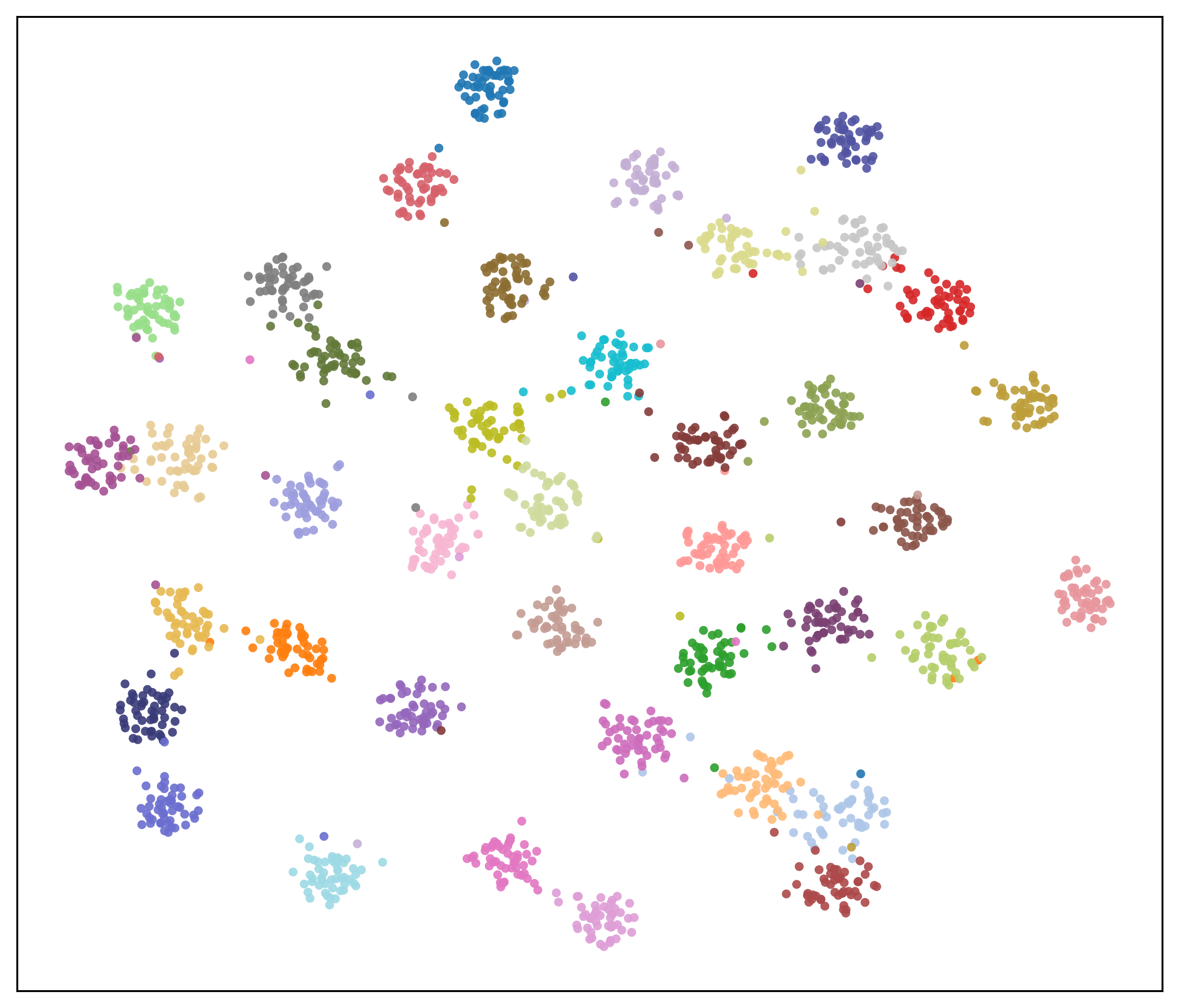}
    \caption{\texttt{\Ours}.}
    \label{fig:txt_aug}
  \end{subfigure}
  \caption{
  t-SNE visualization of feature distributions derived from classifiers constructed under different strategies.
t-SNE visualization of feature distributions obtained from classifiers built with different learning strategies.
  } 
  \label{fig:tsne}
\end{figure}

% To verify the effectiveness of our \texttt{BayesMM}, we visualize the feature distributions learned under different strategies on the ModelNet-C dataset using t-SNE.
% As shown in Figure~\ref{fig:tsne}, the {Cache-based updating} strategy produces entangled and overlapping clusters, indicating that static cache features fail to capture discriminative semantics.
% In contrast, {Geometric distribution learning} yields more compact intra-class patterns and clearer inter-class margins, suggesting that modeling geometric prototypes enhances structural consistency.
% {Textual distribution learning} further aligns visual embeddings with textual priors, producing better-separated clusters and improved semantic coherence.
% Finally, our \texttt{BayesMM} achieves the most distinct and compact manifolds, demonstrating that dynamically integrating geometric and textual evidences through Bayesian model averaging effectively reduces feature ambiguity and yields a more discriminative classifier feature space.
To verify the effectiveness of \texttt{BayesMM}, we visualize feature distributions on ModelNet-C using t-SNE.
As shown in Figure~\ref{fig:tsne}, the {Cache-based updating} strategy produces entangled clusters, showing that static cache features fail to capture discriminative semantics.
{Geometric distribution learning} yields compact intra-class patterns and clear inter-class margins, indicating improved structural consistency.
{Textual distribution learning} further aligns visual embeddings with textual priors, leading to better-separated clusters.
Finally, \texttt{BayesMM} produces the most distinct and compact manifolds, as dynamically integrating geometric and textual evidences through Bayesian averaging reduces feature ambiguity and enhances discrimination.

\section{Conclusion}
% This work presented \texttt{BayesMM}, a training-free Bayesian distribution learning framework for adaptive point cloud recognition under distribution shifts.
% By modeling geometric and textual modalities as Gaussian distributions and integrating them through Bayesian model averaging, BayesMM enables stable, uncertainty-aware adaptation without additional training or auxiliary networks.
% Experiments on corrupted, cross-domain, and large-scale benchmarks demonstrate substantial gains in robustness, stability, and cross-modal generalization over existing cache-based and test-time adaptation methods.
% Visual analyses further show that Bayesian fusion produces compact intra-class clusters and clear inter-class boundaries, yielding a more discriminative feature space for reliable 3D understanding in dynamic environments.

This work presented \texttt{BayesMM}, a training-free dynamic Bayesian distribution learning framework for adaptive point cloud recognition under distribution shifts.
By modeling geometric and textual modalities as Gaussian distributions and integrating them through Bayesian model averaging, BayesMM achieves stable and uncertainty-aware adaptation without additional training or auxiliary networks.
Extensive experiments on corrupted, cross-domain, and large-scale benchmarks show that BayesMM substantially improves robustness, stability, and cross-modal generalization compared with existing cache-based and test-time adaptation methods.
% Visual analyses further confirm that the proposed Bayesian fusion yields compact intra-class clusters and clear inter-class boundaries, forming a more discriminative classifier feature space.
Overall, \texttt{BayesMM} establishes a principled and efficient paradigm for multimodal test-time adaptation toward reliable 3D understanding in dynamic environments.

\section*{Acknowledge}
{
This research is supported by the Local Science and Technology Program (No.2024CSJGG00800).
}
{
    \small
    \bibliographystyle{ieeenat_fullname}
    \bibliography{main}
}

% WARNING: do not forget to delete the supplementary pages from your submission 
\renewcommand{\thetable}{\Alph{table}}
\renewcommand{\thefigure}{\Alph{figure}}
\clearpage
\setcounter{page}{1}
% \maketitlesupplementary
\newpage
\appendix
\section{Implementation Details}\label{appx:implement}

For ULIP~\cite{xue23ulip} and ULIP-2~\cite{xue24ulip2}, we adopt PointBERT~\cite{pointbert} as the point-cloud encoder backbone.
For OpenShape~\cite{liu23openshape}, we follow the official configuration and use its scaled PointBERT variant with 32.1M parameters, as reported in Table~\ref{tab:memory_comparison_openshape} of the Appendix.
For Uni3D, we employ the giant model, whose point encoder contains 1,016.5M parameters. All pretrained weights are obtained directly from their public GitHub repositories.
We report the zero-shot recognition accuracy of these large 3D models as the baselines for comparison, and include Point-Cache~\cite{Point-Cache} for completeness.

To describe point clouds, rather than relying on a single prompt such as “a point cloud object of a \{\verb|class|\}”, we follow ULIP~\cite{xue23ulip} and Point-PRC~\cite{0006KWCY0024} and use 64 diverse text templates. Each template produces a textual description that is encoded into an embedding, and the 64 embeddings are averaged to obtain a class-level representation.

We further incorporate additional semantic detail only in the ModelNet-C evaluation. In this setting, the 64 generic templates are concatenated with 50 GPT-generated class-specific descriptions so that each category is provided with richer and more tailored semantics.
For all subsequent evaluations, we use only the 64 generic templates without any class-specific augmentation. Even with this simplified setting, our method still yields competitive performance, which shows that it does not depend on complex or dataset-specific semantic information. This also indicates that the approach is easy to transfer across datasets without the need to generate separate prompts for each of them.
\section{Derivation of Eq.~\eqref{eq:nu} and Eq.~\eqref{eq:closed_form_update} }\label{appx:derivation}
\subsection{Eq.~\eqref{eq:nu}: Textual prototype MAP}
We derive the MAP estimator of the textual prototype $\boldsymbol{\nu}_c$ in Eq.~\eqref{eq:nu}.
For each class $c$, an LLM produces $M$ paraphrased prompts with embeddings
$\{\mathbf{z}_{c,1}, \dots, \mathbf{z}_{c,M}\}$, from which we compute the empirical mean and scatter:
\begin{equation}
\bar{\mathbf{z}}_c = \frac{1}{M}\sum_{i=1}^M \mathbf{z}_{c,i},
\quad
\mathbf{S}_c = \sum_{i=1}^M
(\mathbf{z}_{c,i} - \bar{\mathbf{z}}_c)
(\mathbf{z}_{c,i} - \bar{\mathbf{z}}_c)^\top.
\end{equation}
We treat $\bar{\mathbf{z}}_c$ as a sufficient statistic summarizing the $M$ paraphrases and
model the latent textual prototype $\boldsymbol{\nu}_c$ as a Gaussian variable with prior
\begin{equation}
p(\boldsymbol{\nu}_c) =
\mathcal{N}\!\big(\boldsymbol{\nu}_c \,\big|\, \mathbf{0},\, \beta^2\mathbf{I}\big),
\label{eq:text_prior}
\end{equation}
where $\beta^2$ controls the prior variance.

Conditioned on $\boldsymbol{\nu}_c$, we assume that the empirical mean $\bar{\mathbf{z}}_c$
is drawn from a Gaussian whose covariance shrinks with the number of paraphrases:
\begin{equation}
p(\bar{\mathbf{z}}_c \mid \boldsymbol{\nu}_c)
= \mathcal{N}\!\big(\bar{\mathbf{z}}_c \,\big|\,
\boldsymbol{\nu}_c,\, \tfrac{1}{M}\mathbf{S}_c\big).
\label{eq:text_likelihood}
\end{equation}
By Bayes' rule, the posterior over $\boldsymbol{\nu}_c$ is
\begin{equation}
p(\boldsymbol{\nu}_c \mid \bar{\mathbf{z}}_c)
\propto
p(\bar{\mathbf{z}}_c \mid \boldsymbol{\nu}_c)\,
p(\boldsymbol{\nu}_c).
\end{equation}

Taking the negative log and omitting constants independent of $\boldsymbol{\nu}_c$ gives
\begin{align}
- \log p(\boldsymbol{\nu}_c \mid \bar{\mathbf{z}}_c)
&= \frac{1}{2\beta^2}
\boldsymbol{\nu}_c^\top \boldsymbol{\nu}_c\nonumber \\
&+ \frac{M}{2}
(\bar{\mathbf{z}}_c - \boldsymbol{\nu}_c)^\top
\mathbf{S}_c^{-1}
(\bar{\mathbf{z}}_c - \boldsymbol{\nu}_c)\nonumber \\
&+ \text{const}.
\label{eq:text_quadratic}
\end{align}
Expanding the second term, we obtain:
\begin{align}
- \log p(\boldsymbol{\nu}_c \mid \bar{\mathbf{z}}_c)
&= \frac{1}{2}\boldsymbol{\nu}_c^\top
\big(\beta^{-2}\mathbf{I}
+ M\mathbf{S}_c^{-1}\big)\boldsymbol{\nu}_c
\nonumber\\
&\quad
- \boldsymbol{\nu}_c^\top
\big(M\mathbf{S}_c^{-1}\bar{\mathbf{z}}_c\big)
+ \text{const},
\end{align}
which matches the canonical Gaussian form in $\boldsymbol{\nu}_c$ with precision
\begin{equation}
\boldsymbol{\Lambda}_c
= \beta^{-2}\mathbf{I}
+ M\mathbf{S}_c^{-1},
\end{equation}
and natural parameter
\begin{equation}
\boldsymbol{\eta}_c
= M\mathbf{S}_c^{-1}\bar{\mathbf{z}}_c.
\end{equation}

Thus the posterior over $\boldsymbol{\nu}_c$ is Gaussian,
\begin{equation}
p(\boldsymbol{\nu}_c \mid \bar{\mathbf{z}}_c)
= \mathcal{N}\!\big(\boldsymbol{\nu}_c \,\big|\,
\boldsymbol{\nu}_c^{\text{MAP}},\, \boldsymbol{\Sigma}_{\nu_c}\big),
\end{equation}
with
\begin{align}
\boldsymbol{\Sigma}_{\nu_c}
&= \boldsymbol{\Lambda}_c^{-1}
= \big(\beta^{-2}\mathbf{I} + M\mathbf{S}_c^{-1}\big)^{-1},
\\[3pt]
\boldsymbol{\nu}_c^{\text{MAP}}
&= \boldsymbol{\Sigma}_{\nu_c}\,\boldsymbol{\eta}_c
= \big(\beta^{-2}\mathbf{I} + M\mathbf{S}_c^{-1}\big)^{-1}
M\mathbf{S}_c^{-1}\bar{\mathbf{z}}_c.
\label{eq:text_map_full}
\end{align}

Since $\mathbf{S}_c$ in Eq.~(3) is an unnormalized scatter matrix, its global scale
can be absorbed into $M$ without changing the relative weighting between the
prior and data terms. Under this convention, simplifying the common scalar factor
yields the compact expression used in the main paper:
\begin{equation}
\boldsymbol{\nu}_c^{\text{MAP}}
=
\big(\beta^{-2}\mathbf{I} + M\mathbf{S}_c^{-1}\big)^{-1}
\mathbf{S}_c^{-1}\bar{\mathbf{z}}_c,
\end{equation}
which gives Eq.~\eqref{eq:nu}.

\subsection{Eq.~\eqref{eq:closed_form_update}: Geometric distribution update}
We next derive the recursive update in Eq.~(12) for a fixed class $c$, omitting the class index when unambiguous. At test-time step $t$, the geometric parameters are:
\begin{equation}
\Theta_t = \{\boldsymbol{\mu}_t, \boldsymbol{\Sigma}_t\},
\end{equation}
and the prior is given by the previous posterior:
\begin{equation}
p(\Theta_t)=p(\Theta_{t-1}\mid \mathbf{x}_{t-1}).
\label{eq:geom_prior_general}
\end{equation}

Under the Gaussian assumptions in Eq.~\eqref{eq:geo_prior} and Eq.~\eqref{eq:geo_likelihood}, the mean evolves according to:
\begin{align}
\boldsymbol{\mu}_t &\sim \mathcal{N}(\boldsymbol{\mu}_{t-1},\, \boldsymbol{\Sigma}_{t-1}), \label{eq:geom_prior_mean}\\
\mathbf{x}_t \mid \boldsymbol{\mu}_t &\sim \mathcal{N}(\boldsymbol{\mu}_t,\, \boldsymbol{\Sigma}). \label{eq:geom_likelihood}
\end{align}
Applying Bayes' rule yields:
\begin{equation}
p(\boldsymbol{\mu}_t \mid \mathbf{x}_t)\propto
p(\mathbf{x}_t\mid \boldsymbol{\mu}_t)\,p(\boldsymbol{\mu}_t).
\end{equation}

Taking the negative log (up to constants independent of $\boldsymbol{\mu}_t$) gives:
\begin{align}
-\log p(\boldsymbol{\mu}_t\mid \mathbf{x}_t)
&=\frac{1}{2}(\boldsymbol{\mu}_t-\boldsymbol{\mu}_{t-1})^\top
\boldsymbol{\Sigma}_{t-1}^{-1}
(\boldsymbol{\mu}_t-\boldsymbol{\mu}_{t-1})
\nonumber\\[-3pt]
&\quad +\frac{1}{2}
(\mathbf{x}_t-\boldsymbol{\mu}_t)^\top
\boldsymbol{\Sigma}^{-1}
(\mathbf{x}_t-\boldsymbol{\mu}_t).
\label{eq:geom_quadratic}
\end{align}

Expanding the quadratic terms leads to:
\begin{align}
-\log p(\boldsymbol{\mu}_t\mid \mathbf{x}_t)
&=\frac{1}{2}\boldsymbol{\mu}_t^\top
(\boldsymbol{\Sigma}_{t-1}^{-1}+\boldsymbol{\Sigma}^{-1})
\boldsymbol{\mu}_t
\nonumber\\[-3pt]
&\quad -\boldsymbol{\mu}_t^\top
\big(\boldsymbol{\Sigma}_{t-1}^{-1}\boldsymbol{\mu}_{t-1}
+\boldsymbol{\Sigma}^{-1}\mathbf{x}_t\big)
+\text{const},
\end{align}
which matches the canonical Gaussian form with precision:
\begin{equation}
\boldsymbol{\Lambda}_t=\boldsymbol{\Sigma}_{t-1}^{-1}+\boldsymbol{\Sigma}^{-1},
\end{equation}
and natural parameter:
\begin{equation}
\boldsymbol{\eta}_t
=\boldsymbol{\Sigma}_{t-1}^{-1}\boldsymbol{\mu}_{t-1}
+\boldsymbol{\Sigma}^{-1}\mathbf{x}_t.
\end{equation}

Thus the posterior is Gaussian with parameters:
\begin{align}
\boldsymbol{\Sigma}_t
&=\boldsymbol{\Lambda}_t^{-1}
=\big(\boldsymbol{\Sigma}_{t-1}^{-1}+\boldsymbol{\Sigma}^{-1}\big)^{-1},
\\[3pt]
\boldsymbol{\mu}_t
&=\boldsymbol{\Sigma}_t\,\boldsymbol{\eta}_t
=\boldsymbol{\Sigma}_t
\big(\boldsymbol{\Sigma}^{-1}\mathbf{x}_t
+\boldsymbol{\Sigma}_{t-1}^{-1}\boldsymbol{\mu}_{t-1}\big).
\end{align}

Restoring the class index and substituting the class-specific covariance $\boldsymbol{\Sigma}^c$, we obtain:
\begin{align}
\boldsymbol{\mu}_t^{\,c}
&=\boldsymbol{\Sigma}_t^{\,c}
\Big[
(\boldsymbol{\Sigma}^c)^{-1}\mathbf{x}_t
+(\boldsymbol{\Sigma}_{t-1}^{\,c})^{-1}\boldsymbol{\mu}_{t-1}^{\,c}
\Big],\\[3pt]
\boldsymbol{\Sigma}_t^{\,c}
&=\Big[
(\boldsymbol{\Sigma}_{t-1}^{\,c})^{-1}
+(\boldsymbol{\Sigma}^c)^{-1}
\Big]^{-1},
\end{align}
which gives Eq.~\eqref{eq:closed_form_update} in the main paper.

% \section{Rationale}
% \label{sec:rationale}
% % 
% Having the supplementary compiled together with the main paper means that:
% % 
% \begin{itemize}
% \item The supplementary can back-reference sections of the main paper, for example, we can refer to \cref{sec:intro};
% \item The main paper can forward reference sub-sections within the supplementary explicitly (e.g. referring to a particular experiment); 
% \item When submitted to arXiv, the supplementary will already included at the end of the paper.
% \end{itemize}
% % 
% To split the supplementary pages from the main paper, you can use \href{https://support.apple.com/en-ca/guide/preview/prvw11793/mac#:~:text=Delete%20a%20page%20from%20a,or%20choose%20Edit%20%3E%20Delete).}{Preview (on macOS)}, \href{https://www.adobe.com/acrobat/how-to/delete-pages-from-pdf.html#:~:text=Choose%20%E2%80%9CTools%E2%80%9D%20%3E%20%E2%80%9COrganize,or%20pages%20from%20the%20file.}{Adobe Acrobat} (on all OSs), as well as \href{https://superuser.com/questions/517986/is-it-possible-to-delete-some-pages-of-a-pdf-document}{command line tools}.
\section{Additional Results}
\noindent{\textbf{Robustness evaluation.}} 
Tables~\ref{tab:s_obj_only_c_robustness}, \ref{tab:s_obj_bg_c_robustness}, and \ref{tab:s_hardest_c_robustness} present the recognition accuracy under different corruption settings. Overall, {\Ours} consistently improves performance over the baseline models (ULIP, ULIP-2, O-Shape, and Uni3D) as well as the Point-Cache variants.

On S-OBJ\_ONLY-C (Table~\ref{tab:s_obj_only_c_robustness}), {\Ours} consistently outperforms Point-Cache, improving the average accuracy by 3.0\% to 4.5\%, and achieves a 4\% to 7\% gain over the original backbone models, demonstrating clear improvements in robustness.
On S-OBJ\_BG-C (Table~\ref{tab:s_obj_bg_c_robustness}), which introduces background clutter, {\Ours} achieves 2.5\% to 3.5\% higher average accuracy than Point-Cache and up to 6\% to 8\% improvement over the original models, indicating strong generalization in more challenging scenes.  
On the most challenging split S-PB\_T50-RS-C (Table~\ref{tab:s_hardest_c_robustness}), {\Ours} increases the average accuracy by 4.5\% to 5.0\% over Point-Cache and by 7\% to 10\% over the original backbones, demonstrating that our approach can effectively handle severe corruptions and partial observations, significantly enhancing model robustness.  
These consistent gains across settings highlight the effectiveness of {\Ours} in improving robustness under diverse and severe corruptions.

\noindent{\textbf{Memory usage and inference throughput.}}
Table~\ref{tab:memory_comparison_openshape} reports the memory consumption across ModelNet-C, Omni3D, and O-LVIS. 
While Point-Cache exhibits comparable or slightly higher memory usage on smaller datasets, its parameter footprint grows substantially as the number of object categories increases, particularly on O-LVIS.
In contrast, {\Ours} maintains a consistently lightweight profile, indicating that the robustness improvements introduced by our method come with only negligible memory overhead.
Table~\ref{tab:throughput_comparison_scan} presents the inference throughput measured on S-OBJ\_ONLY. 
The throughput reduction introduced by {\Ours} remains within 2\% to 5\% across all evaluated backbones, indicating that the additional operations do not significantly affect runtime. 
The resulting throughput decrease remains modest relative to the robustness enhancements achieved by {\Ours}.

\begin{table*}[ht]
    \footnotesize
    \centering
    \caption{Comparison of recognition accuracy on S-OBJ\_ONLY-C, which contains seven types of corruptions. Results are reported at corruption severity level 2. Each clean point cloud contains 1024 points. SONN refers to ScanObjectNN.}
    \label{tab:s_obj_only_c_robustness}
    \begin{tabular}{l c c c c c c c c c}
       \toprule
       \multirow{2}{*}{Method} & \textbf{Original Data} & \multicolumn{7}{c}{\textbf{Corruption Type}} & \multirow{2}{*}{\textbf{Avg.}} \\\cline{3-9}
             & SONN & Add Global & Add Local & Drop Global & Drop Local & Rotate & Scale & Jitter & \\
       \midrule
ULIP~\cite{xue23ulip}
& 49.05 & 31.50 & 34.77 & 51.29 & 38.38 & 48.36 & 44.58 & 36.83 & 41.85 \\

+ Point-Cache (Global)
& \underline{52.15} & \underline{35.80} & \underline{37.01} & 54.39 & 41.82 & 49.74 & 45.09 & \textbf{40.28} & 44.54 \\

+ Point-Cache (Hierarchical)
& 52.15 & 32.01 & \textbf{38.04} & \underline{54.56} & \underline{45.27} & \underline{50.95} & \underline{45.96} & \underline{39.24} & \underline{44.77} \\

+ \textbf{\Ours}
& \textbf{54.04} & \textbf{39.24} & \textbf{38.04} & \textbf{55.42} & \textbf{46.30} & \textbf{52.15} & \textbf{47.68} & 35.11 & \textbf{45.75} \\

       \midrule
ULIP-2~\cite{xue24ulip2} & 42.00 & 40.45 & 41.31 & 37.69 & 30.29 & 38.21 & 44.45 & 22.89 & 37.16 \\ 
\ + Point-Cache (Global) & 48.19 & \underline{49.05} & \underline{46.30} & 45.09 & 37.18 & 41.65 & 44.41 & \underline{25.99} & 42.24 \\
\ + Point-Cache (Hierarchical) & \underline{51.98} & \underline{49.05} & \underline{46.30} & \textbf{48.88} & \underline{40.45} & \textbf{45.78} & \underline{45.09} & \underline{25.99} & \underline{44.19} \\ 
\ + \textbf{\Ours} & \textbf{52.67} & \textbf{54.04} & \textbf{48.02} & \underline{45.78} & \textbf{42.34} & \underline{44.75} & \textbf{45.96} & \textbf{28.57} & \textbf{45.52} \\ 
\midrule
O-Shape~\cite{liu23openshape} 
& 53.18 & 49.91 & 46.30 & 52.15 & 36.66 & 46.64 & 46.82 & 30.81 & 45.31 \\

+ Point-Cache (Global) 
& 56.80 & 56.45 & 51.98 & 54.56 & 40.45 & \underline{51.81} & \underline{49.23} & 37.69 & 49.90 \\

+ Point-Cache (Hierarchical) 
& \underline{58.69} & \underline{59.04} & \underline{53.01} & \underline{55.94} & \underline{41.82} & 51.12 & 48.54 & \underline{39.41} & \underline{50.95} \\

+ \textbf{\Ours} 
& \textbf{61.96} & \textbf{61.10} & \textbf{55.76} & \textbf{59.03} & \textbf{48.54} & \textbf{54.90} & \textbf{53.35} & \textbf{40.96} & \textbf{54.44} \\
\midrule

Uni3D~\cite{zhou24uni3d} 
& 65.58 & 62.65 & 56.45 & 60.07 & 49.40 & 61.62 & 56.11 & 43.55 & 56.93 \\ %

+ Point-Cache (Global) 
& 70.05 & 65.06 & \underline{59.38} & 63.68 & 54.39 & \underline{63.34} & 60.07 & 51.29 & 60.91 \\

+ Point-Cache (Hierarchical) 
& \underline{70.22} & \underline{65.40} & 58.00 & \underline{64.20} & \underline{54.91} & 61.96 & \underline{62.13} & \underline{53.18} & \underline{61.25} \\

+ \textbf{\Ours} 
& \textbf{71.60} & \textbf{69.53} & \textbf{60.06} & \textbf{64.54} & \textbf{60.07} & \textbf{68.33} & \textbf{63.86} & \textbf{52.32} & \textbf{63.79} \\

\bottomrule

    \end{tabular}
\end{table*}

\begin{table*}[ht]
    \footnotesize
    \centering
    \caption{Comparison of recognition accuracy on S-OBJ\_BG-C, which includes seven types of corruptions. Results are reported at corruption severity level 2. Each clean point cloud contains 1024 points.}

    \label{tab:s_obj_bg_c_robustness}
    \begin{tabular}{l c c c c c c c c c}
        \toprule
        \multirow{2}{*}{Method} & \textbf{Original Data} & \multicolumn{7}{c}{\textbf{Corruption Type}} & \multirow{2}{*}{\textbf{Avg.}} \\\cline{3-9}
                & SONN & Add Global & Add Local & Drop Global & Drop Local & Rotate & Scale & Jitter & \\
        \midrule
ULIP~\cite{xue23ulip}
& 45.96 & 27.19 & 25.82 & 45.61 & 34.25 & 40.96 & 40.10 & 30.98 & 36.36 \\

+ Point-Cache (Global)
& 48.88 & 30.46 & \underline{30.46} & \underline{49.05} & 39.59 & \underline{44.92} & \underline{42.17} & \underline{31.84} & \underline{39.68} \\

+ Point-Cache (Hierarchical)
& \underline{49.74} & 28.23 & 30.12 & 48.71 
& \underline{40.45} & 43.55 & 40.28 & \textbf{34.42} & 39.44 \\

+ \textbf{\Ours}
& \textbf{52.67} & \textbf{33.05} & \textbf{34.08} & \textbf{50.43} 
& \textbf{41.65} & \textbf{48.53} & \textbf{45.78} & 31.50 & \textbf{41.81} \\

\midrule

ULIP-2~\cite{xue24ulip2} 
& 48.19 & 40.62 & 38.90 & 39.24 & 32.36 & 41.14 & 42.86 & 21.17 & 38.04 \\ 

+ Point-Cache (Global) 
& 52.50 & 48.19 & 45.09 & 46.82 & 39.07 & 46.64 & 48.02 & \textbf{26}.51 & 44.10 \\

+ Point-Cache (Hierarchical) 
& \underline{54.73} & \textbf{51.64} & \textbf{47.16} & \textbf{50.95} 
& \underline{39.76} & \textbf{53.01} & \textbf{51.81} & 22.72 & \textbf{46.47} \\

+ \textbf{\Ours} 
& \textbf{56.80} & \underline{50.77} & \underline{46.82} & \underline{49.40} 
& \textbf{40.45} & \underline{50.26} & \underline{49.57} & \underline{25.47} & \underline{46.19} \\

\midrule

O-Shape~\cite{liu23openshape} 
& 55.94 & 49.40 & 48.19 & 52.67 & 42.51 & 48.88 & 47.16 & 31.84 & 47.08 \\

+ Point-Cache (Global)
& 59.72 & 57.49 & 51.12 & \underline{59.72} & \underline{48.71} & \underline{56.11} & \underline{54.22} & 35.28 & 52.80 \\

+ Point-Cache (Hierarchical)
& \underline{62.65} & \underline{58.00} & \underline{51.64} & 59.55 
& 47.85 & 54.91 & 53.36 & \underline{36.49} & \underline{53.06} \\

+ \textbf{\Ours}
& \textbf{64.72} & \textbf{60.41} & \textbf{54.90} & \textbf{61.62} 
& \textbf{52.32} & \textbf{60.41} & \textbf{57.14} & \textbf{38.21} & \textbf{55.09} \\

\midrule

Uni3D~\cite{zhou24uni3d} 
& 60.24 & 58.00 & 52.32 & 51.64 & 44.23 & 58.00 & 51.81 & 39.24 & 51.94 \\ %

+ Point-Cache (Global) 
& \underline{63.86} & \textbf{66.27} & \textbf{57.83} & 56.11 & \underline{50.77} & \underline{61.62} & 56.11 & 44.23 & 57.10 \\

+ Point-Cache (Hierarchical) 
& 62.82 & 64.72 & \underline{57.14} & \underline{58.52} 
& 50.43 & 60.93 & \textbf{59.55} & \underline{46.30} & \underline{57.55} \\

+ \textbf{\Ours} 
& \textbf{68.50} & \underline{66.26} & 54.39 & \textbf{60.58} 
& \textbf{55.07} & \textbf{65.23} & \underline{58.86} & \textbf{49.57} & \textbf{59.06} \\
\bottomrule

    \end{tabular}
\end{table*}

\begin{table*}[!ht]
  \footnotesize
  \centering
    \caption{Comparison of corruption generalization on S-PB\_T50-RS-C, the most challenging split of ScanObjectNN. Each clean point cloud is represented by 1024 points. SONN denotes ScanObjectNN.}

  \label{tab:s_hardest_c_robustness}
  \begin{tabular}{l c c c c c c c c c}
     \toprule
     \multirow{2}{*}{Method} & \textbf{Original Data} & \multicolumn{7}{c}{\textbf{Corruption Type}} & \multirow{2}{*}{\textbf{Avg.}} \\\cline{3-9}
           & SONN & Add Global & Add Local & Drop Global & Drop Local & Rotate & Scale & Jitter & \\
     \midrule
ULIP~\cite{xue23ulip} & 29.29 & 19.26 & 18.39 & 30.99 & 23.91 & 27.48 & 26.34 & 21.44 & 24.64 \\
+ Point-Cache (Global) & 32.37 & 22.87 & 20.85 & 33.31 & 27.90 & 30.85 & \underline{28.63} & 24.53 & 27.66 \\ 
+ Point-Cache (Hierarchical) & \underline{32.48} & \underline{23.46} & \underline{22.69} & \underline{34.70} & \underline{31.75} & \underline{33.00} & 28.28 & \underline{25.05} & \underline{28.93} \\ 
+ \Ours & \textbf{40.52} & \textbf{29.53} & \textbf{25.92} & \textbf{39.21} & \textbf{33.59} & \textbf{35.74} & \textbf{32.44} & \textbf{24.67} & \textbf{33.18} \\
\midrule

ULIP-2~\cite{xue24ulip2} & 33.38 & 30.29 & 29.42 & 28.24 & 24.91 & 28.56 & 30.22 & 12.98 & 27.25 \\ 
+ Point-Cache (Global) & 40.28 & \underline{36.40} & 33.80 & 35.39 & 30.88 & 33.66 & 35.01 & 18.36 & 32.97 \\
+ Point-Cache (Hierarchical) & \underline{42.40} & 35.70 & \underline{34.42} & \underline{37.75} & \underline{34.21} & \underline{36.26} & \underline{36.09} & \underline{19.12} & \underline{34.49} \\ 
+ \Ours & \textbf{46.31} & \textbf{41.29} & \textbf{37.82} & \textbf{40.46} & \textbf{34.57} & \textbf{39.73} & \textbf{37.51} & \textbf{16.53} & \textbf{36.78} \\ 
\midrule

O-Shape~\cite{liu23openshape} & 41.12 & 32.41 & 35.60 & 37.80 & 27.34 & 36.61 & 35.22 & 18.88 & 33.12 \\
+ Point-Cache (Global) & 42.16 & 40.32 & 37.58 & 42.02 & 33.76 & 41.53 & \underline{38.24} & 24.12 & 37.47 \\

+ Point-Cache (Hierarchical) & \underline{43.72} & \underline{40.91} & \underline{39.24} & \underline{43.03} & \underline{35.22} & \underline{43.06} & 37.40 & \underline{25.05} & \underline{38.45} \\ 
+ \Ours & \textbf{50.52} & \textbf{49.51} & \textbf{43.64} & \textbf{49.53} & \textbf{41.22} & \textbf{47.11} & \textbf{45.31} & \textbf{30.29} & \textbf{44.39} \\ 
\midrule

Uni3D~\cite{zhou24uni3d} & 46.04 & 48.23 & 37.99 & 36.75 & 31.47 & 44.00 & 37.37 & 28.66 & 38.46 \\
+ Point-Cache (Global) & 50.28 & \underline{52.57} & \underline{42.23} & 42.61 & 36.29 & 47.22 & 39.83 & 33.48 & 43.06 \\

+ Point-Cache (Hierarchical) & \underline{51.13} & 51.67 & 41.88 & \underline{44.59} & \underline{38.79} & \underline{49.03} & \underline{41.05} & \underline{34.70} & \underline{44.10} \\ 
+ \Ours & \textbf{57.04} & \textbf{59.30} & \textbf{45.70} & \textbf{49.10} & \textbf{44.41} & \textbf{53.16} & \textbf{48.37} & \textbf{37.86} & \textbf{49.17} \\ 
\bottomrule

  \end{tabular}
\end{table*}

\begin{table}[!ht]
   \footnotesize
   \centering
   \caption{
   Memory usage (MB) comparison across different datasets. 
Numbers below each dataset name indicate the number of classes. 
Point-Cache denotes the hierarchical variant.
}
   \begin{tabular}{l c c c c}
      \toprule
      \multirow{2}{*}{Method} & ModelNet-C & Omni3D & O-LVIS  & \#Params\\
      & (40) & (216) & (1156) &(M)\\
      \midrule
      ULIP-2 & \textbf{1,556} & \textbf{1,558} & \textbf{1,556} &85.7 \\ 
      \ + Point-Cache & \textbf{1,556} & \textbf{1,558} & 1,566 &85.7 \\
      \ + \textbf{\Ours} & \underline{1,560} & \underline{1,560} & \underline{1,562} &85.7 \\
      \midrule
      OpenShape & \textbf{7,056} & \textbf{7,058} & \textbf{7,116} &2,571.9 \\ 
      \ + Point-Cache & \underline{7,058} & \underline{7,062} & 7,150 &2,571.9 \\
      \ + \textbf{\Ours} & 7,076 & 7,080 & \underline{7,084} &2,571.9 \\
      \bottomrule
   \end{tabular}
   \label{tab:memory_comparison_openshape}
\end{table}

\begin{table}[!ht]
   % \footnotesize
   \tabstyle{4pt}  
   \centering
\caption{%
Inference throughput (samples per second) on S-OBJ\_ONLY.
All experiments are conducted with a batch size of 1 on an RTX 3090 GPU. 
Point-Cache refers to the hierarchical variant of Point-Cache here.
   }
   \begin{tabular}{l c c c c}
      \toprule
      {Method} & ULIP & ULIP2 & OpenShape & Uni3D\\
      \midrule
       Vanilla  & \textbf{11.25} & \textbf{11.25}& \textbf{8.60} & \textbf{7.72} \\ 
      \ + Point-Cache~\cite{Point-Cache}  & \underline{11.17}& \underline{11.17} & \underline{8.57} & \underline{7.62} \\
      \ + \textbf{\Ours}  & 10.90 & 10.91 & 8.28 & 7.41 \\
      \bottomrule
   \end{tabular}
   \label{tab:throughput_comparison_scan}
\end{table}

\newpage
\appendix
\onecolumn

\end{document}